\documentclass[letterpaper]{article}

\usepackage[nonatbib,final]{neurips_2021}

\usepackage[utf8]{inputenc} 
\usepackage[T1]{fontenc}    
\usepackage{hyperref}       
\usepackage{url}            
\usepackage{booktabs}       
\usepackage{amsfonts}       
\usepackage{nicefrac}       
\usepackage{microtype}      
\usepackage{xcolor}         

\usepackage{amsmath,amsfonts,amsthm,amssymb}
\usepackage[noend]{algcompatible}
\usepackage{graphicx,subcaption}
\usepackage{textcomp}
\usepackage{xcolor,soul}
\usepackage[numbers]{natbib}

\usepackage[utf8]{inputenc} 
\usepackage[T1]{fontenc}    
\usepackage{hyperref}       
\usepackage{url}            
\usepackage{booktabs}       
\usepackage{nicefrac}       
\usepackage{microtype}      
\usepackage{algorithm}	    
\usepackage{xcolor}         
\usepackage{todonotes}      
\usepackage{mathtools}
\usepackage{soul}
\usepackage{graphicx}
\usepackage{tikz}
\usepackage{tikz-cd}
\usetikzlibrary{calc}
\usepackage{pgfplots}
\usepgfplotslibrary{groupplots}
\pgfplotsset{compat=1.3}
\usepackage{wrapfig}
\usepackage{cancel}

\usepackage[most]{tcolorbox}
\usetikzlibrary{patterns}
\pgfdeclarepatternformonly{mystrikeout}{\pgfqpoint{-1pt}{-1pt}}{\pgfqpoint{11pt}{11pt}}{\pgfqpoint{10pt}{10pt}}%
{
  \pgfsetlinewidth{0.4pt}
  \pgfpathmoveto{\pgfqpoint{0pt}{0pt}}
  \pgfpathlineto{\pgfqpoint{10.1pt}{10.1pt}}
  \pgfusepath{stroke}
}
\newtcolorbox{tcbstrikeout}{breakable,
 enhanced jigsaw,
 opacityback=0,
 parbox=false,
 boxrule=0mm,
 top=0mm,bottom=0pt,left=0pt,right=0pt,
 boxsep=0pt,
 frame hidden,
 finish={\fill[pattern=mystrikeout] (frame.north west) rectangle (frame.south east);}
}

\usepackage{minitoc}

\newcommand{\paragraphbe}[1]{\paragraph{#1} }

\newtheorem*{rep@theorem}{\rep@title}
\newcommand{\newreptheorem}[2]{%
\newenvironment{rep#1}[1]{%
 \def\rep@title{#2 \ref{##1}}%
 \begin{rep@theorem}}%
 {\end{rep@theorem}}}
\makeatother

\newtheorem*{rep@lemma}{\rep@title}
\newcommand{\newreplemma}[2]{%
\newenvironment{rep#1}[1]{%
 \def\rep@title{#2 \ref{##1}}%
 \begin{rep@lemma}}%
 {\end{rep@lemma}}}
\makeatother

\newtheorem*{rep@corollary}{\rep@title}
\newcommand{\newrepcorollary}[2]{%
\newenvironment{rep#1}[1]{%
 \def\rep@title{#2 \ref{##1}}%
 \begin{rep@corollary}}%
 {\end{rep@corollary}}}
\makeatother

\newtheorem{theorem}{{\bf Theorem}}
\newreptheorem{theorem}{{\bf Theorem}}
\newtheorem{lemma}{{\bf Lemma}}
\newreptheorem{lemma}{{\bf Lemma}}
\newtheorem{corollary}{{\bf Corollary}}
\newreptheorem{corollary}{{\bf Corollary}}
\newtheorem{definition}{{\bf Definition}}[section]

\newcommand{\D}{D} 
\newcommand{\R}{\mathbb{R}} 
\newcommand{\N}{\mathbb{N}} 
\newcommand{\Univ}{\mathcal{X}} 
\newcommand{\Domain}{\Univ^\size} 
\newcommand{\eps}{\varepsilon} 

\newcommand{\dif}[1]{d #1}

\newcommand{\doh}[2]{\frac{\partial #1}{\partial #2}}
\newcommand{\graD}[1]{\nabla #1}
\newcommand{\divR}[1]{\nabla \cdot #1}
\newcommand{\lapL}[1]{\Delta #1}
\newcommand{\hesS}[1]{\nabla^2 #1}
\newcommand{\tra}[1]{\textsf{Tr}\left(#1\right)}
\newcommand{\dotP}[2]{\left\langle#1,#2\right\rangle}
\newcommand{\Evnt}{S}	                             
\newcommand{\ve}{\ensuremath{\mathbf v}}

\newcommand{\ptheta}{\theta}                         

\newcommand{\g}{g}
\newcommand{\sen}[1]{S_{#1}}
\newcommand{\loss}[2]{\ell(#2;\mathbf{#1})}                  
\newcommand{\Loss}{\mathcal{L}}                              
\newcommand{\tnoise}{\sigma}                                 
\newcommand{\dime}{d}		                             
\newcommand{\lip}{L}                                         
\newcommand{\cvx}{\lambda}                                   
\newcommand{\size}{n}                                 	     
\newcommand{\smh}{\beta}                                     
\newcommand{\norm}[1]{\left\lVert#1\right\rVert_2}
\newcommand{\ltwo}{l_2}
\newcommand{\K}{K}					     
\newcommand{\kk}{k}

\newcommand{\rhO}{p}
\newcommand{\nU}{\nu}                                        
\newcommand{\q}{\alpha}
\newcommand{\Ren}[3]{R_{#1}\left(#2\middle\|#3\right)}       
\newcommand{\Fren}[3]{E_{#1}\left(#2\middle\|#3\right)}      
\newcommand{\Gren}[3]{I_{#1}\left(#2\middle\|#3\right)}      
\newcommand{\E}{\mathbb{E}}
\newcommand{\Expec}[2]{\underset{#1}{\E}\left[#2\right]}
\newcommand{\Z}{\mathbf{Z}}
\newcommand{\V}{V}
\newcommand{\Gaus}[2]{\mathcal{N}(#1, #2)}
\newcommand{\Id}{\mathbb{I}_d}
\newcommand{\Frenyi}[3]{E_{#1}\left(#2\middle\|#3\right)}

\newcommand{\btheta}{{\mathbb \theta}}
\newcommand{\Range}{\Theta}
\newcommand{\eqdef}{\stackrel{\text{def}}{=}}

\newcommand{\eqr}[1]{\eqref{#1}}
\newcommand{\alg}[1]{Algorithm~\ref{#1}}

\DeclareMathOperator*{\argmin}{arg\,min}

\newcommand{\maxu}[1]{\underset{#1}{\max}}
\newcommand{\argminu}[1]{\underset{#1}{\argmin}}

\newcommand{\Renyi}[3]{R_{#1}\left(#2\middle\|#3\right)}

\newcommand{\W}[1]{\mathbf{W}_{#1}}
\newcommand{\deltW}[1]{\dif{\W{#1}}}
\newcommand{\smaloss}[2]{\ell(#2;\mathbf{#1})}

\newcommand{\thetaspace}{\mathbb{R}^d}

\newcommand{\proj}[2]{\Pi_{#1}(#2)}
\newcommand{\algo}{\mathcal{A}}
\newcommand{\database}{D}

\newcommand{\GG}[3]{I_{#1}\left(#2\middle\|#3\right)}
\newcommand{\step}{\eta}
\newcommand{\gaussnoise}{\mathbf{Z}}
\newcommand{\gaussvariance}[1]{\mathcal{N}(0,#1\mathbb{I}_d)}
\newcommand{\thet}{\Theta}
\newcommand{\sig}{\sigma}
\newcommand{\x}{\mathbf{x}}

\newcommand{\real}{\mathbb{R}}
\newcommand{\alp}{c}
\newcommand{\lam}{\lambda}
\newcommand{\be}{\beta}
\newcommand{\del}{\delta}

\newcommand{\Pro}[1]{Pr\left[#1\right]}

\newcommand{\Lip}{L}

\newcommand{\Ftheta}[1]{\mathcal{F}_{#1}}
\newcommand{\setdif}{\oplus}
\newcommand{\C}{\mathcal{C}}

\def\polylog{\operatorname{polylog}}

\title{Differential Privacy Dynamics of Langevin Diffusion and Noisy Gradient Descent}

\author{%
  Rishav Chourasia\textsuperscript{*}, Jiayuan Ye\textsuperscript{*}, Reza Shokri\\
  Department of Computer Science,  National University of Singapore\\
  \texttt{\{rishav1, jiayuan, reza\}@comp.nus.edu.sg}
}

\begin{document}

\maketitle

\begingroup\renewcommand\thefootnote{*}
\footnotetext{Equal contribution. Alphabetical Order.}
\endgroup

\doparttoc 
\faketableofcontents 


\begin{abstract}
	What is the information leakage of an iterative randomized learning algorithm about its training data, when the internal state of the algorithm is \emph{private}? How much is the contribution of each specific training epoch to the information leakage through the released model? We study this problem for noisy gradient descent algorithms, and model the \emph{dynamics} of R\'enyi differential privacy loss throughout the training process.  Our analysis traces a provably \emph{tight} bound on the R\'enyi divergence between the pair of probability distributions over parameters of models trained on neighboring datasets.  We prove that the privacy loss converges exponentially fast, for smooth and strongly convex loss functions, which is a significant improvement over composition theorems (which over-estimate the privacy loss by upper-bounding its total value over all intermediate gradient computations). For Lipschitz, smooth, and strongly convex loss functions, we prove optimal utility with a small gradient complexity for noisy gradient descent algorithms.
\end{abstract}


\section{Introduction}
\label{sec:intro}

Machine learning models leak a significant amount of information about their training data, through their parameters and predictions~\cite{shokri2017membership, nasr2019comprehensive, carlini2020extracting}.  Iterative randomized training algorithms can limit this information leakage and bound the differential privacy loss of the learning process~\cite{bassily2014private, abadi2016deep, feldman2018privacy, feldman2020private}.  The strength of this certified defense is determined by an \emph{upper bound} on the (R\'enyi) divergence between the probability distributions of model parameters learned on any pair of neighboring datasets.

The general method to compute the differential privacy bound for gradient perturbation-based learning algorithms is to view the process as a number of (identical) differential privacy mechanisms, and to compute the \emph{composition} of their bounds.  However, this over-estimates the privacy loss of the released model~\cite{jagielski2020auditing, nasr2021adversary}, and results in a loose differential privacy bound.  This is because composition bounds also accounts for the leakage of all intermediate gradient updates, even though only the final model parameters are observable to adversary.  \citet{feldman2018privacy, feldman2020private} address this issue for the privacy analysis of gradient computations over \emph{one single} training epoch, for smooth and convex loss functions.  However, in learning a model over multiple training epochs, such a guarantee is quantitatively similar to composition bounds of privacy amplification by sub-sampling~\cite{feldman2018privacy}. The open challenge, that we tackle in this paper, is to provide an analysis that can tightly bound the privacy loss of the \emph{released model} after~$K$ training epochs, for any~$K$. 

We present a novel analysis for privacy dynamics of noisy gradient descent with smooth and strongly convex loss functions. We construct a pair of continuous-time Langevin diffusion~\cite{sato2014approximation} processes that trace the probability distributions over the model parameters of noisy GD.  Subsequently, we derive differential inequalities bounding the \emph{rate} of privacy loss (worst case R\'enyi divergence between the \textbf{coupled stochastic processes} associated with neighboring datasets) throughout the training process. We then prove an exponentially-fast converging privacy bound for noisy GD: (simplified theorem) Under $1$-strongly convex and $\smh$-smooth loss function~$\loss{\x}{\ptheta}$ with total gradient sensitivity~$1$, the noisy GD Algorithm~\ref{alg:noisygd}, with initial parameter vector~$\ptheta_0\sim\proj{\C}{\mathcal{N}(0,2\sig^2\mathbb{I}_d)}$ and step size~$\eta<\frac{1}{\be}$, satisfies $(\q, \eps)$-R\'enyi DP with
$\eps = O\left(\frac{\q}{\sigma^2\size^2}(1-e^{-\step\frac{\K}{2}})\right)$, where $n$ is the size of the training set.

This guarantee shows that the privacy loss \textbf{converges} exponentially in the number of iterations~$K$, instead of growing proportionally with~$\K$ as in the composition-based analysis of the same algorithms.  Our bound captures a strong privacy amplification due to the dynamics (and convergence) of differential privacy over the noisy gradient descent algorithm with private internal state.  

We analyze the \emph{tightness} of the bound, the \emph{utility} of the randomized algorithm under the computed differential privacy bound, as well as its \emph{gradient complexity} (number of required gradient computations).  We prove the \textbf{tightness guarantee} for our bound by showing that there exist a loss function and neighboring datasets such that the divergence between corresponding model parameter distributions matches our privacy bound. For Lipschitz, smooth, and strongly convex loss functions, we prove that noisy GD achieves \textbf{optimal utility} under differential privacy with an error of order $O(\frac{d}{n^2\eps^2})$, with a \emph{small gradient complexity} of order $O(n\log(n))$. This improves over the prior utility results for noisy SGD algorithms~\cite{bassily2014private}.  Our analysis results in a significantly smaller gradient complexity by a factor of ${\size/\log(\size)}$, and a slightly better utility by a factor of $\polylog(\size)$.

We anticipate that our work will have a positive societal impact, by paving the way for building accurate and privacy preserving machine learning systems for sensitive personal data.


\section{Preliminaries on differential privacy}
\label{sec:short_prelim}

Let $\Univ$ be the data universe, and a dataset~$\D$ contain $n$ records from it: $\D = (\x_1, \x_2, \cdots, \x_\size)\in\mathcal{X}^n$.  We refer to a dataset pair $\D, \D'$ as \emph{neighboring} if they differ in one data record. A measure $\nu$ is said to be absolutely continuous with respect to another measure $\nu'$ on same space (denoted as $\nu \ll \nu'$) if for all measurable set $S$, $\nu(S) = 0$ whenever $\nu'(S) = 0$.

\begin{definition}[\cite{mironov2017renyi} R\'enyi differential privacy]
    \label{def:renyi_diff_priv_short_prelim}
    Let $\q > 1$.  A randomized algorithm $\algo: \Domain \rightarrow \thetaspace$ satisfies \emph{{$(\q, \eps)$-R\'enyi~Differential Privacy} (RDP)}, if for any two neighboring datasets $\D, \D' \in \Domain$, the $\alpha$ R\'enyi divergence $\Ren{\q}{\algo(\D)}{\algo(\D')} \leq \eps$. For a pair of measures $\nu, \nu'$ over the same space with $\nu\ll\nu'$, $\Ren{\q}{\nu}{\nu'}$ is defined as
    \begin{equation}
        \label{eqn:renyi_divergence_short_prelim}
	\Ren{\q}{\nu}{\nu'} = \frac{1}{\q-1}\log\Fren{\q}{\nu}{\nu'}, \quad \text{where}
	\quad \Fren{\q}{\nu}{\nu'}=\int\left(\frac{d\nu}{d\nu'}\right)^\q d\nu'.
    \end{equation}
\end{definition}

We refer to $\Ren{\q}{\algo(\D)}{\algo(\D')}$ also as the \emph{R\'enyi privacy loss} of algorithm~$\algo$ on datasets~$\D, \D'$. An RDP guarantee can be converted to $(\eps, \del)$-DP guarantee~\cite[Proposition~5]{mironov2017renyi}.

\begin{definition}[\cite{vempala2019rapid} R\'enyi information]
    Let $\q > 1$. For any two measures $\nu, \nu'$ over $\thetaspace$ with $\mu\ll\nu'$ and corresponding probability density functions $\rhO, \rhO'$, if $\frac{\rhO(\ptheta)}{\rhO'(\ptheta)}$ is differentiable, the \emph{$\q$-R\'enyi Information} of $\nu$ with respect to $\nu'$ is
    \begin{equation}
        \label{eqn:grenyi_divergence_short_prelim}
        \Gren{\q}{\nu}{\nu'} = \frac{4}{\q^2}\Expec{\ptheta \sim \rhO'}
        {\norm{\graD{\frac{\rhO(\ptheta)^\frac{\q}{2}}{\rhO'(\ptheta)^\frac{\q}{2}}}}^2}
	= \Expec{\ptheta \sim \rhO'}{ \frac{ \rhO(\ptheta)^{\q-2} }{ \rhO'(\ptheta)^{\q-2} } 
	\norm{ \graD{ \frac{\rhO(\ptheta) }{ \rhO'(\ptheta) } } }^2 }.
    \end{equation}
\end{definition}

See the Appendix~\ref{sec:prelim} for a comprehensive presentation of preliminaries.

\section{Privacy analysis of noisy gradient descent}
\label{sec:privacy_combined}

Let $\D = (\x_1, \x_2, \cdots, \x_\size)$ be a dataset of size $\size$ with records taken from a universe $\Univ$. For a given machine learning algorithm, let ${\loss{\x}{\ptheta} : \Univ \times \thetaspace \rightarrow \R}$ be a loss function of a parameter vector $\ptheta \in \C$ on the data point~$\x$, where $\C$ is a closed convex set (can be $\mathbb{R}^d$). 

A generic formulation of the optimization problem to learn the model parameters, is in the form of empirical risk minimization (ERM) with the following objective, where $\Loss_\D(\ptheta)$ is the empirical loss of the model, with parameter vector $\ptheta$, on a dataset $\D$.
\begin{equation}
    \label{eqn:erm_objective}
    \ptheta^* = \argminu{\ptheta \in \C}~\Loss_\D(\ptheta), \quad \text{where} \quad \Loss_\D(\ptheta)=\frac{1}{\size} \sum_{\x \in \D}\loss{\x}{\ptheta}.
\end{equation}

Releasing this optimization output (i.e.,~$\ptheta^*$) can leak information about the dataset $\D$, hence violating data privacy. To mitigate this risk, there exist randomized algorithms 
to ensure that the ($\q$-R\'enyi) privacy loss of the ERM algorithm is upper-bounded by $\eps$, i.e., the algorithm satisfies ${(\q, \eps)\text{-RDP}}$. 

\begin{algorithm}[h]
	\caption{$\algo_{\text{Noisy-GD}}$: Noisy Gradient Descent}
	\label{alg:noisygd}
	\begin{algorithmic}[1]
		\REQUIRE Dataset $\D = (\x_1, \x_2, \cdots, \x_\size)$, loss function $\ell(\theta;\x)$, closed convex set $\C\subseteq\mathbb{R}^d$, learning rate $\step$, noise variance $\tnoise^2$, initial parameter vector $\ptheta_0$.
		\FOR{$\kk = 0, 1, \cdots, \K - 1$}
		\STATE {$g(\ptheta_k;\D)= \sum_{i=1}^n \nabla\ell({\ptheta_k};{\x_i})$}
		\STATE {$\ptheta_{\kk+1} = \proj{\C}{\ptheta_{\kk} - \frac{\step}{n} g(\ptheta_k;\D) + \sqrt{2\step}\Gaus{0}{\tnoise^2\Id}}$} \label{alg:ngd:updatestep}
		\ENDFOR
		\STATE {Output $\ptheta_{\K}$}
	\end{algorithmic}
\end{algorithm}

In this paper, our objective is to analyze privacy loss of \emph{Noisy Gradient Descent} (\alg{alg:noisygd}), which is a randomized ERM algorithm.  Let $\ptheta_k,\ptheta_k'$ be the parameter vectors at the $k$'th iteration of $\algo_{\text{Noisy-GD}}$ on neighboring datasets $\D$ and $\D'$, respectively. We denote by $\thet_{\step\kk}$ and $\thet_{\step\kk}'$ the corresponding random variables that model $\ptheta_k$ and $\ptheta_k'$. We abuse notation to also denote their probability distributions by $\thet_{\step\kk}$ and $\thet_{\step\kk}'$. In this paper, our objective is to model and analyze the {\bf dynamics of differential privacy} of this algorithm. More precisely, we focus on the following.

\begin{enumerate}
	\item Compute an RDP bound (i.e., the worst case R\'enyi divergence~$\Ren{\q}{\thet_\K}{\thet_\K'}$ between the output distributions of two neighboring datasets) for Algorithm~\ref{alg:noisygd}, and analyze its tightness.

	\item Compute the contribution of each iteration 
    to the privacy loss.  As we go from step $\kk = 1$ to~$\K$ in Algorithm~\ref{alg:noisygd}, we investigate how the algorithm's privacy loss changes as it runs the $\kk$'th iteration (computed as~${\Ren{\q}{\thet_{\step\kk}}{\thet_{\step\kk}'} - \Ren{\q}{\thet_{\eta(\kk-1)}}{\thet_{\eta(\kk-1)}'}}$).
\end{enumerate}

In the end, we aim to provide a RDP bound that is tight, thus facilitating optimal utility~\cite{bassily2014private}.
We emphasize that our goal is to construct 
a theoretical framework for analyzing privacy loss of releasing the output $\ptheta_\K$ of the algorithm, 
assuming \emph{private} internal states (i.e., $\ptheta_1, \cdots, \ptheta_{\K-1}$).


\subsection{Tracing diffusion for Noisy GD}
\label{ssec:tracing_diffusion}

To analyze the privacy loss of Noisy GD, which is a \emph{discrete-time stochastic process}, we first interpolate each discrete update from $\ptheta_k$ to $\ptheta_{\kk+1}$ with  a piece-wise continuously differentiable diffusion process. Let $D$ and $D'$ be a pair of arbitrarily chosen neighboring datasets.
Given step-size $\step$ and initial parameter vector~$\ptheta_0=\theta_0'$, 
the respective $\kk$'th discrete updates in \alg{alg:noisygd} on neighboring datasets $D$ and $D'$ are
\begin{equation}
    \label{eqn:discretejump}
    \begin{cases}
    \ptheta_{\kk+1}=\proj{\C}{\ptheta_{\kk}-\step\graD{\Loss_\D(\ptheta_\kk)}+ \sqrt{2\step\tnoise^2}\gaussnoise_\kk}, \\
    
    \ptheta_{\kk+1}'=\proj{\C}{\ptheta_{\kk}'-\step\graD{\Loss_\D(\ptheta_\kk')}+ \sqrt{2\step\tnoise^2}\gaussnoise_\kk},
    \\
    \end{cases} \text{with} \quad \gaussnoise_\kk\sim\mathcal{N}(0,\mathbb{I}_d).
\end{equation}

These two discrete jumps can be interpolated with two stochastic processes $\thet_t$ and $\thet_t'$ over time $\step \kk \leq t\leq\step (\kk+1)$ respectively. At the start of each step, $t=\eta \kk$,  the random variables $\thet_{\eta \kk}$ and $\thet_{\eta \kk}'$ model the distribution of the $\theta_k$ and $\theta_k'$ in the noisy GD processes respectively. During time $\eta k<t<\eta (\kk+1)$, we model the respective gradient updates on $D$ and $D'$ with the following stochastic processes. 
\begin{equation}
\label{eqn:soldiscreteSDE}
\begin{cases}
\thet_{t}=
\Theta_{\eta k}-\eta\cdot U_1(\thet_{\eta k}) - (t-\eta k)\cdot U_2 (\thet_{\eta k}) + \sqrt{2(t-\eta k)\sig^2} \gaussnoise_\kk
\\
\thet_{t}'=
\Theta_{\eta k}'-\eta\cdot U_1(\thet_{\eta k}') + (t-\eta k)\cdot U_2 (\thet_{\eta k}') + \sqrt{2(t-\eta k)\sig^2} \gaussnoise_\kk\\
\end{cases}
\end{equation}
where the vectors $U_1(\theta)=\frac{1}{2}\left(\nabla \Loss_\D(\theta) + \nabla \Loss_{\D'}(\theta)\right)$ and $U_2 (\theta)=\frac{1}{2}\left(\nabla \Loss_\D(\theta) - \nabla \Loss_{\D'}(\theta)\right)$ represent the average and difference between gradients on neighboring datasets $D$ and $D'$ respectively. 

At the end of step, i.e. at $t \rightarrow \eta (k+1)$, we project $\thet_t$ and $\thet_t'$ onto convex set $\C$, and obtain
\begin{equation}
    \label{eqn:discreteprojection}
    \thet_{\eta(\kk+1)}=\Pi_{\mathcal{C}}\left(\lim_{t\rightarrow \eta(\kk+1)^-}\thet_t\right), \thet_{\eta(\kk+1)}'=\Pi_{\mathcal{C}}\left(\lim_{t\rightarrow \eta(\kk+1)^-}\thet_t'\right).
\end{equation}

By plugging \eqr{eqn:soldiscreteSDE} into \eqr{eqn:discreteprojection}, we compute that the projected random variable $\Theta_{\step(\kk+1)}$ and $\Theta_{\step(\kk+1)}'$ have the same distributions as the parameters $\ptheta_{\kk+1}$ and $\ptheta_{\kk+1}'$ at $\kk+1^{\text{th}}$ step of noisy GD respectively.  
Repeating the construction for $\kk=0,\cdots,\K-1$, we define two piece-wise continuous diffusion processes $\{\thet_t\}_{t\geq0}$ and $\{\thet_t'\}_{t\geq0}$
whose distributions at time $t=\step\kk$ are consistent with $\theta_\kk$ and $\theta_{\kk}'$ in the noisy GD processes \eqr{eqn:discretejump}  for any $\kk \in \{0,\cdots,\K-1\}$. 

\begin{definition}[Coupled tracing diffusions] 
\label{def:coupleddiffusion}
	Let $\thet_0=\thet_0'$ be two identically distributed random variables. We refer to the stochastic processes $\{\thet_t\}_{t\geq0}$ and $\{\thet_{t}'\}_{t\geq0}$ that evolve along diffusion processes \eqr{eqn:soldiscreteSDE} in $\eta\kk<t<\eta(\kk+1)$ and undergo projection steps \eqr{eqn:discreteprojection} at the end of step $t=\step(\kk+1)$, as coupled tracing diffusions for noisy GD on neighboring datasets $\D,\D'$.
\end{definition}

The R\'enyi divergence $R_{\alpha}(\thet_{\eta \K}\lVert\thet_{\eta \K}')$ reflects the R\'enyi privacy loss of \alg{alg:noisygd} with $\K$ steps. Conditioned on observing $\ptheta_k$ and $\theta_\kk'$, the processes $\{\Theta_t\}_{\step\kk< t<\step(\kk+1)}$ and $\{\Theta_t'\}_{\step\kk< t<\step(\kk+1)}$ in \eqr{eqn:soldiscreteSDE} are Langevin diffusions along vector fields $ - U_2(\ptheta_{\kk})$ and $U_2(\ptheta_{\kk}')$ respectively, for duration $\step$. Therefore, conditioned on observing $\theta_\kk$ and $\theta_\kk'$, the diffusion processes in ~\eqr{eqn:soldiscreteSDE} have the following stochastic differential equations (SDEs) respectively.

\begin{equation}
    \label{eqn:discreteSDE}
    \dif{\thet_t} = - U_2(\theta_k)\dif{t} + \sqrt{2\tnoise^2}\deltW{t}, \quad \dif{\thet_t}' =  U_2(\theta_k')\dif{t} + \sqrt{2\tnoise^2}\deltW{t},
\end{equation}
where $\deltW{t}\sim\sqrt{dt}\gaussvariance{}$ describe the Wiener processes on $\thetaspace$. Therefore, the conditional probability density functions $\rhO_{t|\step\kk}(\ptheta|\ptheta_k)$ and $\rhO_{t|\step\kk}'(\ptheta|\ptheta_k')$ follow the following Fokker-Planck equation. For brevity, we use $\rhO_{t|\eta \kk}(\ptheta|\ptheta_k)$ and $\rhO_{t|\eta \kk}'(\ptheta|\ptheta_k')$ to represent the conditional probability density function $p(\thet_t = \ptheta | \thet_{\eta k} = \ptheta_k)$ and $p(\thet_t' = \ptheta | \thet_{\eta \kk}' = \ptheta_\kk')$ respectively.
\begin{equation} \label{eqn:tracing_fokker}
\begin{cases}
\doh{\rhO_{t|\step\kk}(\ptheta|\ptheta_\kk)}{t} =  \divR{\left(\rhO_{t|\step\kk}(\ptheta|\ptheta_\kk) U_2(\theta_k)\right)} + \tnoise^2 \lapL{\rhO_{t|\step\kk}(\ptheta|\ptheta_\kk)}\\
\doh{\rhO'_{t|\step\kk}(\ptheta|\ptheta'_\kk)}{t} = - \divR{\left(\rhO'_{t|\step\kk}(\ptheta|\ptheta'_\kk)U_2(\theta_k')\right)} + \tnoise^2 \lapL{\rhO_{t|\step\kk}(\ptheta|\ptheta'_\kk)}
\end{cases}
\end{equation}
By taking expectations over probability density function $p_{\eta k}(\theta_\kk)$ or $p'_{\eta k}(\theta_\kk')$ on both sides of \eqref{eqn:tracing_fokker}, we obtain the partial differential equation that models the evolution of (unconditioned) probability density function $\rhO_t(\ptheta)$ and $\rhO_t'(\ptheta)$ in the coupled tracing diffusions.
\begin{lemma}
    \label{lem:GDupdate_fokker}
    For coupled tracing diffusion processes \eqr{eqn:soldiscreteSDE} in time $\step\kk < t < \step(\kk+1)$, the equivalent Fokker-Planck equations are
    \begin{equation} \label{eqn:GDupdate_fokker}
    \begin{cases}
    \negthickspace\doh{\rhO_t(\ptheta)}{t} =  \divR{(\rhO_t(\ptheta)V_t(\theta))} + \tnoise^2\lapL{\rhO_t(\ptheta)}\\
    \negthickspace\doh{\rhO_t'(\ptheta)}{t} =  \divR{(\rhO_t'(\ptheta)V'_t(\theta))} + \tnoise^2\lapL{\rhO_t'(\ptheta)},\\
    \end{cases}
    \end{equation}
    where $V_t(\theta) = \Expec{\ptheta_\kk \sim \rhO_{\step\kk|t}}
        {U_2(\ptheta_\kk)|\ptheta}$ and $V'_t(\theta) = \Expec{\ptheta_\kk' \sim \rhO'_{\step\kk|t}}
        {- U_2(\ptheta_\kk)|\ptheta}$ are time-dependent vector fields on $\mathbb{R}^d$, and ${U_2 (\theta)=\frac{1}{2}\left[\nabla \Loss_D(\theta) - \nabla \Loss_{D'}(\theta)\right]}$ is the difference between gradients on neighboring datasets.
\end{lemma}

By this density evolution equation, we model the noisy gradient descent updates with coupled tracing diffusions. The tracing diffusion process is similar to Langevin diffusion. Therefore, we first study the privacy dynamics in coupled tracing (Langevin) diffusions.

\subsection{Privacy erosion in tracing (Langevin) diffusion}
\label{ssec:privacy_diffusion}
The R\'enyi divergence (privacy loss) $\Ren{\q}{\thet_t}{\thet_t'}$ between coupled tracing diffusion processes increases over time, as the vector fields $V_t, V_t'$ underlying two processes are different.  We refer to this phenomenon as {\bf privacy erosion}. This increase is determined by the amount of change in the probability density functions for coupled tracing diffusions, characterized by the Fokker-Planck equations \eqr{eqn:GDupdate_fokker} for diffusions under different vector fields.

Using equation~\eqr{eqn:GDupdate_fokker}, we compute a bound on the rate (partial derivative) of $\Ren{\q}{\thet_t}{\thet_t'}$ over time in the following lemma, to model privacy erosion between two different diffusion processes. We refer to \emph{coupled diffusions} as respective diffusion processes under different vector fields $V_t$ and $V_t'$. 

\begin{lemma}[Rate of R\'enyi privacy loss]
    \label{lem:marginalrenyi}
    Let $V_t$ and $V_t'$ be two vector fields on $\thetaspace$ corresponding to a pair of arbitrarily chosen neighboring datasets $\D$ and $\D'$ with $\maxu{\ptheta \in \thetaspace} \norm{V_t(\ptheta) - V_t'(\ptheta)} \leq \sen{v}$ for all $t\geq 0$. Then, for corresponding coupled diffusions $\{\thet_t\}_{t\geq0}$ and $\{\thet_t'\}_{t\geq0}$ under vector fields $V_t$ and $V_t'$ and noise variance $\tnoise^2$, the R\'enyi privacy loss rate at any $t\geq0$ is upper bounded by
    \begin{equation} \label{eqn:marginalrenyi}
		\doh{\Ren{\q}{\thet_{t}}{\thet_{t}'}}{t}
		\leq 
	    \frac{1}{\gamma}\frac{\q \sen{v}^2}{4\tnoise^2}
	    -
	    (1-\gamma)\tnoise^2\q \frac{\Gren{\q}{\thet_{t}}{\thet_{t}'}}{\Fren{\q}{\thet_{t}}{\thet_{t}'}}.
    \end{equation}
    where $\gamma>0$ is a tuning parameter that we can fix arbitrarily according to our need.
\end{lemma}

Although this lemma bounds the R\'enyi privacy loss rate, the term $\GG{\q}{\thet_{t}}{\thet_{t}'}$ depends on unknown distributions $\thet_{t}, \thet_{t}'$, and is intractable to compute.  Even with explicit expressions for distributions $\thet_{t}, \thet_{t}'$, the calculation would involve integration in $\thetaspace$ which is computationally prohibitive for large $d$. Note that, however, the ratio $I_\q/E_\q$ is always positive by definition.  Therefore, the R\'enyi divergence (privacy loss) rate in \eqr{eqn:marginalrenyi} is bounded by its first component (a constant) given any fixed $\q$. 

\begin{theorem}[Linear R\'enyi divergence bound]
    \label{thm:linearRDP}
	Let $V_t$ and $V_t'$ be two vector fields on $\thetaspace$, with $\maxu{\ptheta \in \thetaspace} \norm{V_t(\ptheta) - V_t'(\ptheta)} \leq \sen{v}$ for all $t\geq 0$. Then, the coupled diffusions under vector fields $V_t$ and $V_t'$ with noise variance $\sig^2$ for time $T$ has $\alpha$-R\'enyi divergence bounded by $\eps=\frac{\alpha \sen{v}^2 T}{4\sig^2}$.
\end{theorem}

When the vector fields are $V_t = - \graD{\Loss_\D}$ and $V_t' = - \graD{\Loss_{\D'}}$, the coupled diffusions follow Langevin diffusion. By definition~\ref{def:gradient_sensitivity} of total gradient sensitivity, ${\maxu{\ptheta \in \thetaspace} \norm{\graD{\Loss_\D}(\ptheta) - \graD{\Loss_{\D'}(\ptheta)}} \leq \frac{\sen{g}}{\size}}$. Therefore, this na\"ive privacy analysis gives linear RDP guarantee for Langevin diffusion, which resembles the moment accountant analysis~\cite{abadi2016deep}. However, a tighter bound of R\'enyi privacy loss is possible with finer control of the ratio ${\GG{\q}{\thet_{t}}{\thet_{t}'}}/{\Fren{\q}{\thet_t}{\thet_t'}}$, which by definition depends on the likelihood ratio between $\thet_t$ and $\thet_t'$, thus is connected with R\'enyi privacy loss itself. When this ratio grows, 
the R\'enyi privacy loss rate decreases, thus slowing down privacy loss accumulation, and leading to tighter privacy bound. 

\paragraphbe{Controlling R\'enyi privacy loss rate under isoperimetry}
We control the ${I_\q}/{E_\q}$ term in lemma~\ref{lem:marginalrenyi} by making an \emph{isoperimetric} assumption known as \emph{log-Sobolev inequality}~\cite{bakry2013analysis}, described as follows.

\begin{definition}[\cite{gross1975logarithmic}~Log-Sobolev Inequality ($\alp$-LSI)]
    Distribution of a random variable $\thet$ on $\thetaspace$ satisfies \emph{logarithmic Sobolev inequality} with parameter $\alp > 0$, i.e. it is $\alp$-LSI, if for all functions $f$ in the function set ${\Ftheta{\thet}=\{f:\thetaspace\rightarrow\mathbb{R}\vert \nabla f\ \text{is continuous, and}\ \mathbb{E}(f(\thet)^2)<\infty\}}$, we have
    \begin{equation} \label{eqn:lsi_standard}
        \mathbb{E}[f(\thet)^2 \log f(\thet)^2]
        - \mathbb{E}[f(\thet)^2] \log \mathbb{E}[f(\thet)^2] 
	\leq \frac{2}{\alp} \mathbb{E}[\norm{\graD{f(\thet)}}^2].
    \end{equation}
\end{definition}
LSI was introduced by \citet{gross1975logarithmic} as a necessary and sufficient condition for rapid convergence of a diffusion processes. Recently, \citet{vempala2019rapid} showed that this isoperimetry condition is sufficient for rapid convergence of Langevin diffusion in R\'enyi divergence. Under LSI, they provide the following useful lower bound on ${I_\q}/{E_\q}$ for an arbitrary pair of distributions.
\begin{lemma}[\cite{vempala2019rapid} $\alp$-LSI in terms of Rényi Divergence]
    \label{lem:RDinLSI}
    Suppose $\thet_t, \thet_t' \in \thetaspace$ are random variables such that the density ratio between distributions of $\thet_t$ and $\thet_t'$ lies in $\Ftheta{\thet_t'}$.
    Then for any $\q \geq 1$, 
    \begin{align} \label{eqn:RDinLSI}
        \Ren{\q}{\thet_t}{\thet_t'} 
        + 
        \q(\q - 1) \doh{\Ren{\q}{\thet_t}{\thet_t'}}{\q}  
        \leq 
        \frac{\q^2}{2\alp} \frac{\Gren{\q}{\thet_t}{\thet_t'}}{\Fren{\q}{\thet_t}{\thet_t'}},
    \end{align}
    if and only if distribution of $\thet_t'$ satisfies $\alp$-LSI.
\end{lemma}
Note that $\doh{\Ren{\q}{\thet_t}{\thet'_t}}{\q}$ is always positive, as ${\Ren{\q}{\thet_t}{\thet'_t}}$ monotonically increases with $\q>1$~\cite{mironov2017renyi}. This lemma shows that ${\Gren{\q}{\thet_t}{\thet_t'}}/{\Fren{\q}{\thet_t}{\thet_t'}}$ grows monotonically with the R\'enyi privacy loss $\Renyi{\q}{\thet_t}{\thet_t'}$. By Lemma~\ref{lem:marginalrenyi}, this implies a throttling privacy loss rate as privacy loss accumulates.Combining Lemma~\ref{lem:marginalrenyi} and Lemma~\ref{lem:RDinLSI}, we therefore model the \textbf{dynamics for R\'enyi privacy loss under $\alp$-LSI} with the following PDE, which describes the relation between privacy loss, its changes over time, and its change over R\'enyi parameter $\q$. For brevity, let $R(\q,t)$ represent $\Ren{\q}{\thet_t}{\thet_t'}$.
\begin{equation} \label{eqn:EvolvePDE}
     \doh{R(\q,t)}{t} 
     \leq
     \frac{1}{\gamma}\frac{\q S_v^2}{4\tnoise^2}
     - 2(1-\gamma)\tnoise^2\alp \left[ \frac{R(\q,t)}{\q}
     + (\q-1)\doh{R(\q,t)}{\q} \right]
\end{equation}

The initial privacy loss $R(\q,0)=0$, as $\thet_0 = \thet_0'$. 
The solution for this PDE increases with time $t\geq 0$, and models the erosion of R\'enyi privacy loss in coupled tracing diffusions $\Theta_t$ and $\Theta_t'$.


\subsection{Privacy guarantee for Noisy GD}
\label{ssec:privacy_noisygd}

We now use the privacy dynamics \eqref{eqn:EvolvePDE} of coupled tracing diffusions to analyze the privacy dynamics for noisy GD. We first bound the difference between the underlying vector fields $V_t$ and $V_t'$ for coupled tracing diffusions for noisy GD on neighboring datasets $D$ and $D'$.

\begin{lemma}
    \label{lem:new_delayed_loss_diff}
    Let $\loss{\x}{\ptheta}$ be a loss function on closed convex set $\C$, with a finite total gradient sensitivity $\sen{\g}$. Let $\{\thet_t\}_{t\geq0}$ and $\{\thet_t'\}_{t\geq0}$ be the coupled tracing diffusions for noisy GD on neighboring datasets $\D, \D' \in \Domain$, under loss ${\smaloss{\x}{\ptheta}}$ and noise variance $\tnoise^2$. Then the difference between underlying vector fields $V_t$ and $V_t'$ for coupled tracing diffusions is bounded by
    \begin{equation}
        \label{eqn:newdiscrete_error}
        \max_{\theta\in\mathbb{R}^d}\Vert V_t(\theta) - V_t'(\theta)\Vert_2 \leq \frac{S_g}{n},
    \end{equation}
    where $V_t(\theta)$ and $V'_t(\theta)$ are time-dependent vector fields on $\mathbb{R}^d$, defined in Lemma~\ref{lem:GDupdate_fokker}.
\end{lemma}

We then substitute $S_v$ in PDE~\eqr{eqn:EvolvePDE} with $S_g/n$, and compute the following PDE modelling R\'enyi privacy loss dynamics of tracing diffusion at $\step\kk<t<\step(\kk+1)$, under $c$-LSI condition.
\begin{equation} \label{eqn:evolvePDEdiscrete}
    \doh{R(\q,t)}{t} 
    \leq
    \frac{1}{\gamma}\frac{\q \sen{\g}^2}{4\tnoise^2\size^2}
    - 2(1-\gamma)\sig^2\alp \left[ \frac{R(\q,t)}{\q}     
    + (\q-1)\frac{\partial R(\q,t)}{\partial\q} \right]
\end{equation}
We solve this PDE under $\gamma=\frac{1}{2}$ for each time piece, and combine multiple pieces by seeing projection as privacy-preserving post-processing step. We derive the RDP guarantee for the Noisy GD algorithm.
\begin{theorem}[RDP for noisy GD under $\alp$-LSI]
    \label{thm:RDbounddiscrete}
    Let $\{\thet_t\}_{t\geq0}$ and $\{\thet_t'\}_{t\geq0}$ be the tracing diffusion for $\algo_{\text{Noisy-GD}}$ on neighboring datasets $\database$ and $\database'$, under noise variance $\sig^2$ and loss function $\ell(\theta;\x)$. Let $\loss{\x}{\ptheta}$ be a loss function on closed convex set $\C\subseteq \mathbb{R}^d$, with a finite total gradient sensitivity $\sen{\g}$. 
    If for any neighboring datasets $\database$ and $\database'$, the corresponding coupled tracing diffusions $\thet_{t}$ and $\thet_t'$ satisfy $\alp$-LSI throughout ${0\leq t\leq \step\K}$, then $\algo_{\text{Noisy-GD}}$ satisfies $(\q, \eps)$ R\'enyi Differential Privacy for
    \begin{equation} \label{eqn:RDbounddiscrete}
        \eps = \frac{\q \sen{\g}^2}{2\alp\sig^4\size^2}(1-e^{-\sig^2\alp\step\K}).
    \end{equation}
\end{theorem}

This theorem offers a strong converging privacy guarantee, on the condition that $c$-LSI is satisfied throughout the Noisy GD process. We then analyze the LSI constant $c$ for given Noisy GD process.

\paragraphbe{Isoperimetry constants for noisy GD}


\pgfplotsset{
    /pgf/declare function={
        dp_dynamics(\k,\a,\l,\e,\g,\s,\n,\b) = \a*\g^2/(\l*\s^2*\n^2)*(1-exp(-\l*\e*\k/2);
    }
}

\pgfplotsset{
    /pgf/declare function={
        baseline(\k,\a,\l,\e,\g,\s,\n) = \e*\a*\g^2/(4*\n^2*\s^2)*\k;
    }
}

\begin{figure}
    \centering 
    \begin{tikzpicture}[scale=0.8]

        \begin{axis}[
        no markers,
        samples=50,
        xmin=0,
        ymin=0,
        ymax=0.06,
        xlabel=$K$,
        ylabel={$\eps$ in $(\q, \eps)$-R\'enyi Differential Privacy},
        xmax=500,clip=false,axis y line*=left,axis x line*=bottom,
        legend style={at={(1.4,0.9)},anchor=west}]
            \addplot[thick,black,dashed,domain=0:245] {baseline(x,30,2,0.02,4,0.02,5000)};
            \addlegendentry{Composition}
            \addplot[thick,black,domain=0:500] {dp_dynamics(x,30,1,0.02,4,0.02,5000,4)};
            \addlegendentry{Our bound}
            \addplot[thick,purple,dashed,domain=0:370] {baseline(x,20,2,0.02,4,0.02,5000)};
            \addplot[thick,blue,dashed,domain=0:500] {baseline(x,10,2,0.02,4,0.02,5000)};
            \addplot[thick,violet,domain=0:500] {dp_dynamics(x,20,1,0.02,4,0.02,5000,4)};
            \addplot[thick,blue,domain=0:500] {dp_dynamics(x,10,1,0.02,4,0.02,5000,4)};
            \addplot[thick,orange,domain=0:500] {dp_dynamics(x,10,2,0.02,4,0.02,5000,4)};
            \addplot[thick,red,domain=0:500] {dp_dynamics(x,10,4,0.02,4,0.02,5000,4)};
            \node[anchor=west] at (axis cs: 500,{dp_dynamics(500,30,1,0.02,4,0.02,5000,4)}) {$\q=30,\lambda=1$};
            \node[anchor=west] at (axis cs: 500,{dp_dynamics(500,20,1,0.02,4,0.02,5000,4)}) {$\q=20,\lambda=1$};
            \node[anchor=west] at (axis cs: 500,{dp_dynamics(500,10,1,0.02,4,0.02,5000,4)}) {$\q=10,\lambda=1$};
            \node[anchor=west] at (axis cs: 500,{dp_dynamics(500,10,2,0.02,4,0.02,5000,4)}) {$\q=10,\lambda=2$};
            \node[anchor=west] at (axis cs: 500,{dp_dynamics(500,10,4,0.02,4,0.02,5000,4)}) {$\q=10,\lambda=4$};
        \end{axis}
        
    \end{tikzpicture}
    \caption{Rényi privacy loss of noisy GD over $K$ iterations, quantified using our DP Dynamics Analysis. We show $\eps$ in the $(\q,\eps)$-RDP guarantee derived by Corollary~\ref{cor:main_theorem} (bold lines), and the Baseline composition analysis (dashed lines). 
    We evaluate under the following setting: RDP order ${\alpha\in\{10,20,30\}}$;\ \  $\lam$-strongly convex loss function with $\lam \in \{1, 2, 4\}$; $\be$-smooth loss function with $\be=4$; finite $\ell_2$-sensitivity $S_g$ for total gradient with $S_g=4$; size of the data set $\size=5000$; step-size $\step=0.02$; noise standard deviation $\sig=0.02$. The expressions for computing the privacy loss are: our analysis: $\eps=\frac{\q S_g^2}{\lambda\sig^2 n^2}\cdot \left(1-e^{-\lam\step\K/2}\right)$; and Baseline composition-based analysis (derived by moment accountant~\cite{abadi2016deep} with details in Appendix~\ref{sec:appendixtightness}): ${\eps=\frac{\q S_g^2}{4n^2\sig^2}\cdot\step\K}$.} \label{fig_ourbound}
\end{figure}
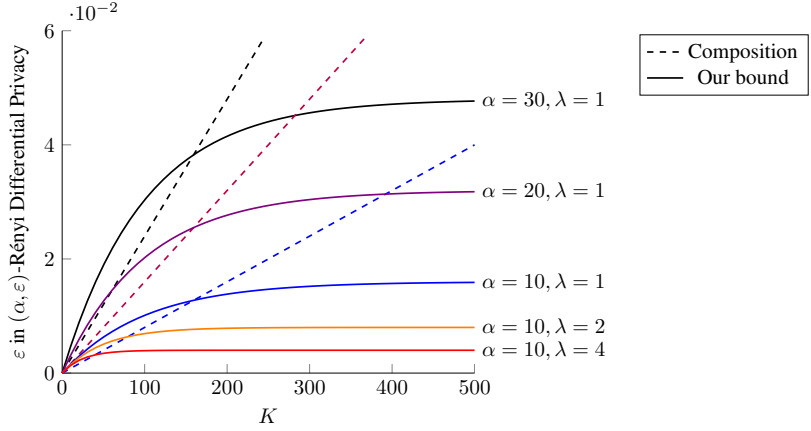

When the loss function is strongly convex and smooth, we prove that tracing diffusion of noisy GD satisfies LSI. This is because the gradient descent update is Lipschitz under smooth loss, and the Gaussian noise preserves LSI, as discussed in Appendix~\ref{ssec:deferred_privacy_noisygd}.

\begin{lemma}[LSI for noisy GD]
    \label{lem:iso_convex}
    If loss function $\loss{\x}{\ptheta}$ is $\cvx$-strongly convex and $\smh$-smooth over a closed convex set $\C$, the step-size is ${\step < \frac{1}{\smh}}$, and initial distribution is $\thet_{0} \sim \proj{\C}{\Gaus{0}{\frac{2\tnoise^2}{\cvx}\Id}}$, then the coupled tracing diffusion processes $\{\thet_t\}_{t\geq 0}$ and $\{\thet_t'\}_{t\geq 0}$ for noisy GD on any neighboring datasets $\D$ and $\D'$ satisfy $\alp$-LSI for any $t\geq0$ with  $\alp=\frac{\lam}{2\sig^2}$. 
\end{lemma}

Using the LSI constant proved by this lemma, we immediately prove the following RDP bound for noisy GD on Lipschitz smooth strongly convex loss, as a corollary of Theorem~\ref{thm:RDbounddiscrete}.

\begin{corollary}[Privacy Guarantee for noisy GD]
    \label{cor:main_theorem}
    Let $\loss{\x}{\ptheta}$ be a $\lam$-strongly convex, and $\be$-smooth loss function on closed convex set $\C$, with a finite total gradient sensitivity $\sen{\g}$, then the noisy gradient descent algorithm (Algorithm~\ref{alg:noisygd}) with start parameter $\ptheta_0 \sim \proj{\C}{\mathcal{N}({0},{\frac{2\tnoise^2}{\lam}\Id})}$, and step-size $\step < \frac{1}{\be}$, satisfies $(\q, \eps)$ R\'enyi Differential Privacy with
    \begin{equation*}
        \eps = \frac{\q S_g^2}{\lam \sigma^2\size^2}(1-e^{-\lam\step\K/2}).
    \end{equation*}
\end{corollary}

This privacy bound has quadratic dependence on the total gradient sensitivity $S_g$, which is upper bounded by $S_g\leq 2\Lip$ for $\Lip$-Lipschitz loss functions. The smoothness condition $\beta$ restricts the step-size and ensures Lipschitz gradient mapping, thus facilitating LSI by Lemma~\ref{lem:iso_convex}. Figure~\ref{fig_ourbound} demonstrates how this RDP guarantee for noisy GD converges with the number of iterations $\K$. Through y-axis, we show the $\eps$ guaranteed for noisy GD under various R\'enyi divergence orders $\alp$ and strong convexity constant $\lam$.
The RDP order $\q$ linearly scales the asymptotic guarantee, but does not affect the convergence rate of RDP guarantee. However, the strong convexity parameter $\lam$ positively affects the asymptotic guarantee as well as the convergence rate; the larger the strong convexity parameter $\lam$ is, the stronger the asymptotic RDP guarantee and the faster the convergence.


\section{Tightness analysis: a lower bound on privacy loss of noisy GD}
\label{ssec:tightness}

Differential privacy guarantees reflect a \emph{bound} on privacy loss on an algorithm; thus, it is very crucial to also have an analysis of their tightness (i.e., how close they are to the exact privacy loss).  We prove that our RDP guarantee in Theorem~\ref{thm:RDbounddiscrete} is tight.  To this end, we construct an instance of the ERM optimization problem, for which we show that the R\'enyi privacy loss of the noisy GD algorithm grows at an order matching our guarantee in Theorem~\ref{thm:RDbounddiscrete}.  

It is very challenging to lower bound the the exact R\'enyi privacy loss ${\Ren{\q}{\thet_{\K\step}}{\thet_{\K\step}'}}$ in general. This might require having an explicit expression for the probability distribution over the last-iterate parameters~$\ptheta_\kk$.  Computing a close-form expression is, however, feasible when the loss gradients are linear.  This is due to the fact that, after a sequence of linear transformations and Gaussian noise additions, the parameters follow a Gaussian distribution. Therefore, we construct such an ERM objective, compute the exact privacy loss, and prove the following lower bound.

\begin{theorem}[Lower bound on RDP of $\algo_{\text{Noisy-GD}}$]
    \label{thm:tightness}
    There exist two neighboring datasets ${\D, \D' \in \Domain}$, a start parameter $\ptheta_0$, and a smooth loss function $\loss{\x}{\ptheta}$ on unconstrained convex set~$\C=\mathbb{R}^d$, with a finite total gradient sensitivity~$\sen{\g}$, such that for any step-size $\step<1$, noise variance $\tnoise^2>0$, and iteration $\K\in\mathbb{N}$, the privacy loss of $\algo_{\text{Noisy-GD}}$ on $\D, \D'$ is lower-bounded by
    \begin{equation}
        \Ren{\q}{\thet_{\step\K}}{\thet_{\step\K}'} \geq \frac{\q S_g^2}{4\sig^2 n^2} \left( 1 - e^{-\step\K} \right).
    \end{equation}
\end{theorem}

We prove this lower bound using the $\ell_2$-squared norm loss as ERM objective: $\min_{\theta\in\mathbb{R}^d}\sum_{i=1}^n\frac{\lVert\theta-\x_i\rVert_2^2}{n}$. We assume bounded data domain 
s.t. the gradient 
has finite sensitivity. 
With start parameter $\theta_0 = 0^d$, the $k^{\text{th}}$ step parameter $\theta_k$ is distributed as Gaussian with mean ${\mu_k=\step\bar{\x}\sum_{i=0}^{k-1}(1-\step)^i}$ and variance ${\sigma_k^2=\frac{2\eta\sigma^2}{\size^2}\sum_{i=0}^{k-1}(1-\step)^{2i}}$ 
in each dimension, where $\bar{x}=\sum_{i=1}^n\x_i/n$ is the empirical dataset mean. 
We explicitly compute the privacy loss 
at any step $\K$, which is lower bounded by ${\frac{\q S_g^2}{4\sig^2n^2}(1-e^{-\eta\K})}$.

Meanwhile, Corollary \ref{cor:main_theorem} gives our RDP upper bound $\epsilon=\frac{\alpha S_g^2}{\sig^2n^2}\left(1-e^{-\eta \K}\right)$ for this same ERM objective. This upper bound matches the lower bound at every step $K$, up to a small constant of $4$.

Moreover, Theorem~\ref{thm:RDbounddiscrete} facilitates a smaller RDP upper bound than Corollary~\ref{cor:main_theorem} by bounding the LSI constant throughout Noisy GD exactly. For squared-norm loss function, Theorem~\ref{thm:RDbounddiscrete} gives the following tighter RDP upper-bound for Noisy GD, because all intermediate Gaussian distributions satisfy $c$-LSI with $c=\frac{2-\eta}{2\sig^2}$, as proved in Appendix~\ref{sec:appendixtightness}.

\begin{corollary}[RDP guarantee of $\algo_{\text{Noisy-GD}}$ on $\ell_2$-norm squared loss]
    \label{cor:tightness}
    For any two neighboring datasets $\D,\D'\in\Domain$, start parameter $\theta_0$, step-size $\step$, noise variance $\tnoise^2$, and $\K \in \N$, if the loss function is $\ell_2$-norm squared loss ${\ell(\theta;\x)=\frac{1}{2}\norm{\theta-\x}^2}$ on unconstrained convex set $\C=\mathbb{R}^d$, with a finite total gradient sensitivity $S_g$, the privacy loss of $\algo_{\text{Noisy-GD}}$ on $\D, \D'$ is upper-bounded by
    \begin{equation}
        \Ren{\q}{\thet_{\step\K}}{\thet_{\step\K}'} \leq \frac{\q S_g^2 }{(2-\step)\sig^2n^2}(1-e^{-\frac{2-\step}{2}\step\K}).
    \end{equation}
\end{corollary}

This RDP guarantee converges fast to $O(\frac{\alpha S_g^2}{\sig^2n^2})$, which matches the lower bound at every step $\K$, up to a constant of~$\frac{4}{2-\eta}\approx 2$. This immediately shows tightness of our converging RDP guarantee \emph{throughout} the training process, for a converging noisy GD algorithm. A different approach is to completely ignore the dynamics of differential privacy, and instead analyze privacy \emph{only at the convergence time} (or when the algorithm is near convergence). \citet{wang2015privacy, minami2016differential} show that sampling from the Gibbs posterior distribution ${\nU(\ptheta) \propto e^{-\Loss_\D(\ptheta)/\tnoise^2}}$ for bounded $\Loss_\D$ satisfies differential privacy. However, sampling exactly from the Gibbs distribution is difficult~\cite{bolstad2009understanding}. Thus, \citet{minami2016differential, ganesh2020faster} extend the DP guarantees of Gibbs posterior distribution to gradient-descent based samplers such as Unadjusted Langevin Algorithm (ULA) that can sample from distributions arbitrarily close to Gibbs distribution after a sufficient number of iterations~$\K$ with extremely small step-size~$\step$. \citet{minami2016differential} compute the distance to convergence in total variation, and \citet{ganesh2020faster} improve the prior bound by measuring the distance in R\'enyi divergence (building on the rapid convergence results of \citet{vempala2019rapid}). The latter results in a better gradient complexity $\Omega(\size \dime)$, which however still grows with model dimension~$\dime$. In comparison, our DP guarantees are unaffected by parameter dimension $\dime$, which in practice can be much larger than the dataset size~$\size$.

In contrast, composition-based privacy bound grows linearly as training proceeds, as shown in Figure~\ref{fig_ourbound}. When the number of iterations $\K$ is small, however, composition-based bound grows at the same rate with the lower bound, as discussed in Appendix~\ref{sec:appendixtightness}. Therefore, to conclude whether our RDP guarantee is superior to composition-based bound, we need to understand the number of iterations noisy GD needs, to achieve optimal utility. We discuss this in the following section.

\section{Utility analysis for noisy gradient descent} \label{ssec:utility}

The randomness, required for satisfying differential privacy, can adversely affect the utility of the trained model.  The standard way to measure the utility of a randomized ERM algorithm (for example, $\algo_{\text{Noisy-GD}}$) is to quantify its worst case \emph{excess empirical risk}, which is defined as 
\begin{equation}
	\label{eqn:excess_emp_risk}
	\max_{\D \in \Domain}\mathbb{E}[\Loss_{\database}(\ptheta) - \Loss_{\database}(\ptheta^*)],
\end{equation}
where $\ptheta$ is the output of the randomized algorithm $\algo_{\text{Noisy-GD}}$ on $\D$, $\ptheta^*$ is the solution to the standard (no privacy) ERM~\eqref{eqn:erm_objective}, and the expectation is computed over the randomness of the algorithm.

We provide the \emph{optimal} excess empirical risk (utility) of noisy GD algorithm under $(\q, \eps')$-RDP constraint. The notion of \emph{optimality} for utility is defined as the smallest upper-bound for excess empirical risk that can be guaranteed under $(\q, \eps')$-RDP constraint by tuning the algorithm's hyper-parameters (such as the noise variance $\tnoise^2$ and the number of iterations $\K$).  We focus here on smooth and strongly convex loss functions with a finite total gradient sensitivity.
	
\begin{lemma}[Excess empirical risk for smooth and strongly convex loss]
	\label{lem:pointwise_conv}
	For $\Lip$-Lipschitz, $\lambda$-strongly convex and $\beta$-smooth loss function $\ell(\theta;\x)$ over a closed convex set $\C\subseteq\mathbb{R}^d$, step-size $\eta\leq \frac{\lambda}{2\beta^2}$, and start parameter $\theta_0\sim\proj{\C}{\mathcal{N}(0,\frac{2\sig^2}{\lam}\Id)}$, the excess empirical risk of Algorithm~\ref{alg:noisygd} is bounded by
	\begin{equation}
		\mathbb{E}[\Loss_{\database}(\theta_{K}) - \Loss_{\database}(\theta^*)] \leq  \frac{2\beta L^2}{\lambda^2}e^{-\lambda\eta K} +\frac{2\beta d\sigma^2}{\lambda},
	\end{equation}
	where $\theta^*$ is the minimizer of $\Loss_\D({\theta})$ in the relative interior of convex set $\mathcal{C}$, and $d$ is the dimension of parameter.
\end{lemma}

This lemma shows decreasing excess empirical risk for noisy GD algorithm under Lipschitz smooth strongly convex loss function as the number of iterations $\K$ increases. The utility is determined by $\K$ and the noise variance $\sig^2$, which are constrained under $(\q,\eps')$-RDP. 
Using our tight RDP guarantee in Corollary~\ref{cor:main_theorem}, we prove optimal utility for noisy GD.

\begin{table}[t!]
    \caption{Utility comparison with the prior $(\eps,\del)$-DP ERM algorithms. We assume $1$-Lipschitz, $\be$-smooth and $\lam$-strongly convex loss.
    Size of input dataset is $\size$, and dimension of parameter vector $\theta$ is $d$. For objective perturbation, we assume $\eps \geq \frac{\smh}{2\lam}$, and loss is twice differentiable. For our result, we assume $\eps \leq 2\log(1/\del)$. The lower bound is $\Omega\left(\min\left\{1, \frac{d}{\eps^2 \size^2}\right\}\right)$~\cite{bassily2014private}. We ignore numerical constants and multiplicative dependence on $\log(1/\delta)$.}
    \label{tab:utility-table}
    \centering
    \small 
    \vspace{0.15cm}
    \begin{tabular}{|l|l|l|l|}
        \hline 
	 & Method & Utility Upper Bound & Gradient complexity \\[0.14cm]
        \hline
        \citet{bassily2014private} 
         & Noisy SGD & $O(\frac{d \log^2(n) }{\lambda n^2\epsilon^2})$ & $n^2$ \\[0.14cm]
        \hline
        \citet{wang2017differentially} 
        & DP-SVRG  & $O(\frac{d \log(n)}{\lambda n^2\epsilon^2})$ & $O\left((n+\frac{\beta}{\lam})\log(\frac{\lam n^2\eps^2}{d})\right)$ \\[0.14cm]
        \hline
        \citet{zhang2017efficient} 
        & Output Perturbation  & $O(\frac{\be d}{\lambda^2 n^2\epsilon^2})$ & $O(\frac{\be}{\lam}n\log(\frac{n^2\eps^2}{d}))$ \\[0.14cm]
        \hline
        \citet{kifer2012private} 
        & Objective Perturbation & $O(\frac{d}{\lambda n^2\epsilon^2})$ & NA \\[0.14cm]
        \hline
        This Paper 
        &  Noisy GD & $O(\frac{\beta d}{\lambda^2 n^2\epsilon^2})$ & $O(\frac{\beta^2}{\lambda^2}n\log\left(\frac{n^2\eps^2}{d}\right))$  \\[0.14cm]
        \hline
    \end{tabular}
\end{table}

\begin{theorem}[Upper bound for $(\q,\eps')$-RDP and $(\eps,\del)$-DP Noisy GD]
    \label{thm:projectedexcessrisk}
	For Lipschitz smooth strongly convex loss function $\ell(\theta;\x)$ on a bounded closed convex set $\mathcal{C}\subseteq\thetaspace$, and dataset $\D\in\Domain$ of size $n$, if the step-size $\eta=\frac{\lambda}{2\beta^2}$ and the initial parameter $\theta_0\sim\proj{\C}{\mathcal{N}(0,\frac{2\sig^2}{\lambda}\mathbb{I}_d)}$, then the noisy GD Algorithm~\ref{alg:noisygd} is $(\alpha,\eps')$-R\'enyi differentially private, where $\alpha>1$ and $\eps'>0$, and satisfies
	\begin{equation}
		\mathbb{E}[\Loss_D(\theta_{K^*})-\Loss_D(\theta^*)]=O(\frac{\alpha\beta d L^2}{\eps'\lam^2\size^2}),
	\end{equation}
	by setting noise variance $\sig^2=\frac{4\alpha \Lip^2}{\lambda\eps' n^2}$, and number of updates $K^*=\frac{2\beta^2}{\lam^2}\log(\frac{n^2\eps'}{\alpha d})$.
	
	Equivalently, for $\eps\leq 2\log(1/\delta)$ and $\delta>0$, Algorithm~\ref{alg:noisygd} is $(\eps,\delta)$-differentially private, and satisfies
	\begin{equation}
		\label{eqn:projectedexcessrisk}
		\mathbb{E}[\Loss_D(\theta_{K^*})-\Loss_D(\theta^*)]=O(\frac{\beta d L^2\log(1/\delta)}{\epsilon^2\lam^2\size^2}),
	\end{equation}
	by setting noise variance $\sig^2=\frac{8\Lip^2 (\eps+2\log(1/\del))}{\lambda\eps^2 n^2}$, and number of updates $K^*=\frac{2\beta^2}{\lam^2}\log(\frac{n^2\eps^2}{4\log(1/\delta) d})$.
\end{theorem}

Our algorithm achieves this utility guarantee with $O(\frac{\beta^2}{\lambda^2}n\log\left(\frac{\eps^2n^2}{d}\right))$ gradient computations of $\nabla\ell(\theta;\x)$, which is faster than noisy SGD algorithm~\cite{bassily2014private} with a factor of $n$. However, we additionally assume smoothness for the loss function. Our gradient complexity also matches that of other efficient gradient perturbation and output perturbation methods~\cite{wang2017differentially,zhang2017efficient}, as shown in Table~\ref{tab:utility-table}.

This utility matches the following theoretical lower bound in \citet{bassily2014private} for the best attainable utility of $(\epsilon,\delta)$-differentially private algorithms on Lipschitz smooth strongly convex loss functions.

\begin{theorem}[\cite{bassily2014private} Lower bound for $(\eps,\del)$-DP algorithms]
    \label{thm:utility_lower_bound}
    Let $\size, \dime \in \N$, $\eps > 0$, and $\delta=o(\frac{1}{n})$. For every $(\eps, \del)$-differentially private algorithm $\algo$ (whose output is denoted by $\theta^{priv}$), there is a dataset $\D \in \Domain$ such that, with probability at least $1/3$ (over the algorithm random coins), we must have
    \begin{equation}
	\Loss_\D(\ptheta^{priv}) - \Loss_\D(\ptheta^*) = \Omega\left(\min\left\{1, \frac{d}{\eps^2 \size^2}\right\}\right),
    \end{equation}
    where $\ptheta^*$ minimizes a constructed $1$-Lipschitz, $1$-strongly convex objective $\mathcal{L}_D(\theta)$ over convex set $\C$.
\end{theorem}

Our utility matches this lower bound upto the constant factor $\log(1/\delta)$, when assuming $\frac{\beta}{\lambda^2}=O(1)$.
This improves upon the previous gradient perturbation methods~\cite{bassily2014private,wang2018differentially} by a factor of $\log(n)$, and matches the utility of previously know optimal ERM algorithm for Lipschitz smooth strongly convex loss functions, such as objective perturbation~\cite{chaudhuri2011differentially,kifer2012private} and output perturbation~\cite{zhang2017efficient}.

\paragraphbe{Utility gain from tight privacy guarantee} As shown in Table~\ref{tab:utility-table}, our utility guarantee for noisy GD is logarithmically better than that for noisy SGD in ~\citet{bassily2014private}, although the two algorithms are extremely similar. This is because we use our tight RDP guarantee, while \citet{bassily2014private} use a composition-based privacy bound. More specifically, noisy SGD needs $n^2$ iterations to achieve the optimal utility, as shown in Table~\ref{tab:utility-table}. This number of iterations is large enough for the composition-based privacy bound to grow above our RDP guarantee, thus leaving room for improving privacy utility trade-off, as we further discuss in Appendix~\ref{sec:projectionproof}. This concludes that our tight privacy guarantee enables providing a superior privacy-utility trade-off, for Lipschitz, strongly convex, and smooth loss functions. 

Our algorithm also has significantly smaller gradient complexity than noisy SGD~\cite{bassily2014private}, for strongly convex loss functions, by a factor of~${n}/{\log n}$. We use a (moderately large) constant step-size, thus achieving fast convergence to optimal utility. However, noisy SGD~\cite{bassily2014private} uses a decreasing step-size, thus requiring more iterations to reach optimal utility.



\section{Conclusions}
\label{sec:conclusions}

We have developed a novel theoretical framework for analyzing the dynamics of privacy loss for noisy gradient descent algorithms.  Our theoretical results show that by hiding the internal state of the training algorithm (over many iterations over the whole data), we can tightly analyze the rate of information leakage throughout training, and derive a bound that is significantly tighter than that of composition-based approaches. 

\paragraphbe{Future Work.} Our main result is a tight privacy guarantee for Noisy GD on smooth and strongly convex loss functions.  The assumptions are very similar to that of the prior work on privacy amplification by iteration~\cite{feldman2018privacy}, and have obvious advantages in enabling the tightness and utility analysis. However, the remaining open challenge is to extend this analysis to non-smooth and non-convex loss functions, and stochastic gradient updates, which are used notably in deep learning. 

\section*{Acknowledgements and Funding}

The authors would like to thank Joe P. Chen, Hedong Zhang and anonymous reviewers for helpful discussions on the earlier versions of this paper. 

This research is supported in part by Intel Faculty Award (within the www.private-ai.org center), Huawei, Google Faculty Award, VMWare Early Career Faculty Award, and the National Research Foundation, Singapore under its Strategic Capability Research Centres Funding Initiative (any opinions, findings and conclusions or recommendations expressed in this material are those of the authors and do not reflect the views of National Research Foundation, Singapore).

\bibliographystyle{plainnat}
\bibliography{reference}

\newpage
\appendix
\noptcrule
\part{Appendix}
\parttoc
\newpage




\section{Table of Notations}

\begin{table}[H]
    \caption{Symbol reference}
    \label{tab:symbol}
    \begin{minipage}{\columnwidth}
        \begin{center}
            \begin{tabular}{ll}
                \toprule
                \textbf{Symbol}                             & \textbf{Meaning}                                                                                  \\
                \hline
                $d$                                         & Dimension of model parameters.                                                                    \\
                $\mathbb{R}^d$                              & Unconstrained model parameter space of dimension $d$.                                             \\
                $\C$                                        & A closed convex Model parameter set $\C\subseteq \thetaspace$ for convex optimization.            \\
                $\Pi_{\mathcal{C}}(\theta)$                 & Projection of $\theta$ to $\mathcal{C}$.                                                          \\
                $\mathcal{X},\mathcal{X}^n$                               & Data universe and Domain of all datasets of size $n$.                                                                                    \\
                $n$                                         & Dataset size.                                                                                     \\
                $D, D'$                 & Neighbouring Dataset of size $n$.                                                                 \\
                $\mathbf{x}_i$                              & $i$-th data point in dataset $D$.                                                       \\
                $\ell(\theta;\mathbf{x})$                   & Risk of parameter $\theta$ w.r.t data point $\mathbf{x}$.                                         \\
                $\mathcal{L}_{D}(\theta)$         & Emprical risk optimization objective.                                                             \\
                ${U_1}(\theta)$         & Average between gradients on neighbouring datasets $D$ and $D'$.                                                             \\
                ${U_2}(\theta)$         & Half of difference between gradients on neighbouring datasets $D$ and $D'$.                                                             \\
                $g(\theta; D)$                             & Sum of risk gradients at $\theta$ for all data points in $\D$.                                    \\
                $V_t, V_t'$                             & Time-variable vector fields on $\mathbb R^d$.                                    \\
                $S_g$                                       & $\ell_2$-sensitivity of total loss gradient $g(\theta; D)$                                       \\
                $S_v$                                       & maximum $\ell_2$ distance between $V_t$ and $V_t'$ for all $t>0$.     \\
                $\theta^*$                                  & Parameter minimizing the empirical risk $\mathcal{L}_{\mathcal{D}}(\theta)$.                      \\
                $\mathcal{L}$                               & Potential function for Langevin diffusion.                                                        \\
                $\mathbf{W}_t$                              & Standard Brownian motion aka. Wiener process.                                                     \\
                $\alpha$                                    & R\'enyi differential privacy order.                                                                \\
                $\delta$                                    & Probability of uncontrolled breach in standard DP.                                                \\
                $\varepsilon$                               & R\'enyi or standard DP privacy parameter.                                                             \\
                $\mathcal{A}$                               & Randomized algorithm.                                                                             \\
                $\nu^{\phantom{'}}, \nu'$                                     & Two probability measures.                                 \\
                $p^{\phantom{'}}, p'$                                     & Two Probability densities over parameter space $\mathbb{R}^d$.                                 \\
                $\Theta, \Theta'$                           & Two random variables distributed as $p, p'$ respectively.                                                    \\
                $\sigma^2$                                  & Noise variance in noisy GD and Langevin diffusion.                                                \\
                $\mathbb{I}_d$                              & $d$-dimensional identity matrix.                                                                  \\
                $\mathcal{N}(0,\mathbb{I}_d)$               & Standard gaussion distribution with dimension $d$.                                                                   \\
                $Z, Z_1, Z_2, \cdots$                       & Random variables taken from $\mathcal{N}(0, \mathbb{I}_d)$.                                       \\
                $\eta$                                      & Step size of updates in noisy GD.                                                                 \\
                $\lambda$                                   & Strong convexity parameter of risk function.                                                      \\
                $\beta$                                     & Smoothness parameter of risk function.                                                            \\
                $B$                                         & Bound on range of risk function.                                                                  \\
                $L$                                         & Lipschitzness parameter of risk function.                                                         \\
                $K,k$                                         & Number of update steps and intermediate step index in noisy GD.                                                               \\
                $\theta^{\phantom{'}}_k, \theta'_k$         & Parameter at step $k$ of noisy GD on $\mathcal{D}, \mathcal{D}'$.                                 \\
                $T,t$                                         & Termination time and intermediate time stamp for diffusion.                                              \\
                $\Theta^{\phantom{'}}_t, \Theta'_t$         & Model parameter random variable at time $t$ of diffusion on $\mathcal{D}, \mathcal{D}'$.                     \\
                $p^{\phantom{'}}_t, p'_t$                   & Probability densities or random variables $\Theta^{\phantom{'}}_t, \Theta'_t$                                           \\
                $p^{\phantom{'}}_{t_1,t_2}$                 & Joint density between diffusion random variables $(\Theta^{\phantom{'}}_{t_1},\Theta^{\phantom{'}}_{t_2})$.                         \\
                $p'_{t_1,t_2}$                              & Joint density between diffusion random variables $(\Theta'_{t_1},\Theta'_{t_2})$.                       \\
                $p^{\phantom{'}}_{t_1|t_2}(\theta|\theta_{t_2})$& Conditional density for $\Theta^{\phantom{'}}_{t_1}$ given $\Theta^{\phantom{'}}_{t_2}=\theta^{\phantom{'}}_{t_2}$.                    \\
                $p'_{t_1|t_2}(\theta|\theta_{t_2})$         & Conditional density for $\Theta_{t_1}'$ given $\Theta_{t_2}'=\theta^{\phantom{'}}_{t_2}$.                  \\
                $R_{\alpha}(\Theta_t||\Theta'_t)$               & R\'enyi divergence of distribution of $\Theta^{\phantom{'}}_t$ w.r.t $\Theta'_t$.                                               \\
		$E_{\alpha}(\Theta_t||\Theta'_t)$               & $\alpha^{\text{th}}$ moment of likelihood ratio r.v. between $\Theta^{\phantom{'}}_t, \Theta'_t$.                                  \\
                $I_{\alpha}(\Theta_t||\Theta'_t)$               & R\'enyi Information of distribution of $\Theta^{\phantom{'}}_t$ w.r.t $\Theta'_t$.                                              \\
                $c$                                         & Constant in Log sobolev inequality.                                                               \\
                 $\ll$                                       & Absolute continuity with respect to measure.                                                          \\
                \bottomrule
            \end{tabular}
        \end{center}
        \bigskip 
    \end{minipage}
\end{table}


\section{Preliminaries}
\label{sec:prelim}

\subsection{Divergence measures}
\label{ssec:prob_measures}

A measure $\nu$ is said to be absolutely continuous with respect to another measure $\nu'$ on same space (denoted as $\nu \ll \nu'$) if for all measurable set $S$, $\nu(S) = 0$ whenever $\nu'(S) = 0$.

\begin{definition}[$\q$-R\'enyi Divergence]
    For $\q > 1$, and any two measures $\nu, \nu'$ with $\nu \ll \nu'$, the \emph{$\q$-R\'enyi Divergence $\Ren{\q}{.}{.}$} of $\nu$ with respect to $\nu'$ is defined as
    \begin{equation}
        \label{eqn:renyi_divergence}
		\Ren{\q}{\nu}{\nu'} = \frac{1}{\q-1}\log\Fren{\q}{\nu}{\nu'},
    \end{equation}
    where $\Fren{\q}{\nu}{\nu'}$ is defined as:
    \begin{equation}
        \Fren{\q}{\nu}{\nu'}=\int\left(\frac{\dif{\nu}}{\dif{\nu'}}\right)^{\q}\dif{\nu'},
    \end{equation}
    
    Additionally, if $\nu$ and $\nu'$ are absolutely continuous with Lebesgue measures on $\mathbb{R}^d$ (i.e. they are continuous distributions on $\thetaspace$) with densities $\rhO$ and $\rhO'$ respectively, $\Fren{\q}{\nu}{\nu'}$ is same as
    \begin{equation}
        \label{eqn:frenyi_divergence}
        \Fren{\q}{\nu}{\nu'}=\Expec{\ptheta \sim \rhO'}{\frac{\rhO(\ptheta)^\q}{\rhO'(\ptheta)^\q}}.
    \end{equation}
\end{definition}

As an example, the $\q$-R\'enyi divergence between two Gaussian distributions centered at $\mu, \mu' \in \thetaspace$, with covariance matrix $\tnoise^2 \Id$ is $\frac{ \q \norm{\mu - \mu'}^2 }{ 2\tnoise^2 }$~\cite[Proposition~7]{mironov2017renyi}.

\begin{definition}[R\'enyi information~\cite{vempala2019rapid}]
    Let $1 < \q < \infty$. For any two measures $\nu, \nu'$ with  $\nu \ll \nu'$, if the Radon-Nikodym derivative $\frac{\dif{\nu}}{\dif{\nu'}}$ is differentiable, the \emph{$\q$-R\'enyi Information} $\Gren{\q}{\cdot}{\cdot}$ of $\nu$ with respect to $\nu'$ is

    \begin{equation}
        \Gren{\q}{\nu}{\nu'} = \int\left(\frac{\dif{\nu}}{\dif{\nu'}}\right)^\q \norm{\graD{\log \frac{\dif{\nu}}{\dif{\nu'}}}}^2 \dif{\nu'}.
    \end{equation}
    Additionally, if $\nu$ and $\nu'$ are absolutely continuous with Lebesgue measures (i.e. they are continuous distributions on $\thetaspace$) with densities $\rhO$ and $\rhO'$ respectively, $\Gren{\q}{\nu}{\nu'}$ is same as
    \begin{equation}
        \label{eqn:grenyi_divergence}
        \Gren{\q}{\nu}{\nu'} = \frac{4}{\q^2}\Expec{\ptheta \sim \rhO'}
        {\norm{\graD{\frac{\rhO(\ptheta)^\frac{\q}{2}}{\rhO'(\ptheta)^\frac{\q}{2}}}}^2} = \Expec{\ptheta \sim \rhO'}{ \frac{ \rhO(\ptheta)^{\q-2} }{ \rhO'(\ptheta)^{\q-2} } \norm{ \graD{ \frac{\rhO(\ptheta) }{ \rhO'(\ptheta) } } }^2 }.
    \end{equation}
\end{definition}

\subsection{Differential privacy}
\label{ssec:diff_privacy}

Let $\Univ$ be a data universe. Let a dataset be a vector of $n$ records from $\mathcal{X}$: $\D = (\x_1, \x_2, \cdots, \x_\size)\in\mathcal{X}^n$.

\begin{definition}[Neighboring datasets]
    Two datasets $\D$ and $\D'$ are \emph{neighboring}, denoted by $\D\sim\D'$, if $|D| = |D'|$, and they differ in exactly one data record, i.e., $\lvert \D\setdif\D' \rvert=2$.
\end{definition}

\begin{definition}[Differential drivacy~\cite{dwork2014algorithmic}]
    \label{def:diff_priv}
    A randomized algorithm $\algo: \Domain \rightarrow \thetaspace$ satisfies \emph{{$(\eps, \del)$-differential~privacy (DP)}} if for any two neighboring datasets $\D, \D' \in \Domain$, and for all sets $\Evnt \in \thetaspace$, 
    \begin{equation}
        \label{eq:diff_priv}
        \Pro{\algo(\D) \in \Evnt} \leq e^\eps \Pro{\algo(\D') \in \Evnt} + \del.
    \end{equation}
\end{definition}

\begin{definition}[R\'enyi differential privacy~\cite{mironov2017renyi}]
    \label{def:renyi_diff_priv}
    Let $\q > 1$.  A randomized algorithm $\algo: \Domain \rightarrow \thetaspace$ satisfies \emph{{$(\q, \eps)$-R\'enyi~Differential Privacy}(RDP)}, if for any two neighboring datasets $\D, \D' \in \Domain$:
    \begin{equation}
        \Ren{\q}{\algo(\D)}{\algo(\D')} \leq \eps.
    \end{equation}
\end{definition}

In this paper, we mainly use R\'enyi differential privacy notion to analyze the privacy loss of algorithms. We refer to $\Ren{\q}{\algo(\D)}{\algo(\D')}$ as the \emph{R\'enyi privacy loss} of algorithm~$\algo$ on datasets~$\D, \D'$.

\begin{theorem}[RDP composition theorem~{{\cite[Proposition~1]{mironov2017renyi}}}]
    \label{thm:RDP_compos}
    Let ${\algo_1: \Domain \rightarrow \thetaspace}$ and ${\algo_2: \thetaspace \times \Domain \rightarrow \thetaspace}$ be two randomized algorithms that satisfy $(\q, \eps_1)$ and $(\q, \eps_2)$-RDP, respectively. The composed algorithm defined as $\algo(\D) = ( \algo_1(\D),  \algo_2(\D))$ satisfies $(\q, \eps_1 + \eps_2)$-R\'enyi~DP.
\end{theorem}

An RDP guarantee can be converted to a DP guarantee as per the following theorem.
\begin{theorem}[DP Conversion~{{\cite[Proposition~3]{mironov2017renyi}}}]
    \label{thm:RDP_to_DP}
    If a randomized algorithm $\algo: \Domain \rightarrow \thetaspace$ satisfies $(\q, \eps)$-RDP, then it also satisfies the standard $(\eps + \frac{\log 1/\del}{\q - 1}, \del)$-DP guarantee for any $0 < \del < 1$.
\end{theorem}

\subsection{Langevin diffusion}
\label{ssec:langevin_diff}

We focus on the Langevin diffusion process in $\mathbb{R}^d$ with noise variance $\sig^2$, described by the following stochastic differential equation (SDE).
\begin{equation} \label{eqn:SDELangevin}
    \dif{\thet_t} = -\graD{\Loss(\thet_t)}\dif{t} + \sqrt{2\tnoise^2}\deltW{t},
\end{equation}
where $d\W{t}=\W{t+dt}-\W{t}\sim\sqrt{\dif{t}}\gaussvariance{}$ characterizes the $d$-dimensional Wiener process.

The joint effect of this drag force (i.e. $-\graD{\Loss}$) and Brownian fluctuations on the probability density $\rhO_t$ of position random variable $\thet_t$ is characterized through the Fokker-Planck equation~\cite{fokker1914mittlere},
\begin{equation} \label{eqn:fokkerplanck_prelim}
    \doh{\rhO_{t}(\ptheta)}{t} = \divR{\left(\rhO_{t}(\ptheta)\graD{\Loss(\ptheta)}\right)} +
    \tnoise^2\lapL{\rhO_{t}(\ptheta)},
\end{equation}
which describes the rate of change in probability density at any position $\ptheta \in \thetaspace$. It's important to point out that Fokker-Planck equation isn't a property of Langevin diffusion, but rather a general equation quantifying the distributional change under \emph{any} drag force in presence of Brownian fluctuations.

Under mild regularity conditions on the potential $\Loss(\ptheta)$, this diffusion process has a stationary distribution $\nU$, given by the solution to ${\doh{\rhO_{t}(\ptheta)}{t} = 0}$, which is the following Gibbs distribution.
\begin{equation} \label{eqn:gibbs}
    \nu(\ptheta)=\frac{1}{\V}e^{-\Loss(\ptheta)/\tnoise^2},\ \text{where} \
    \V=\int_{\thetaspace}e^{-\Loss(\ptheta)/\tnoise^2}\dif{\ptheta}.
\end{equation}

\subsection{Loss function properties}
\label{ssec:func_properties}

For any data record $\x\in\Univ$, a loss function $\smaloss{\x}{\ptheta}: \C\rightarrow\real$ on a closed convex set $\C$ maps parameter $\ptheta \in \C\subseteq\thetaspace$ to a real value. Let $\graD{\loss{\x}{\ptheta}}$ be its loss gradient vector with respect to $\ptheta$.

\begin{definition}[Lipschitz continuity]
    \label{def:lipschitz}
    Function $\loss{\x}{\ptheta}$ is \emph{$\lip$-Lipschitz continuous} if for all $\ptheta, \ptheta' \in \C$ and $\x \in \Univ$,
    \begin{equation} \label{eqn:lipschitz}
        \lvert \loss{\x}{\ptheta} - \loss{\ptheta'}{\x} \lvert \leq \lip \norm{\ptheta - \ptheta'}.
    \end{equation}
\end{definition}

\begin{definition}[Smoothness]
    \label{def:smooth}
    Differentiable function $\loss{\x}{\ptheta}$ is \emph{$\smh$-smooth} over $\C$ if for all $\ptheta, \ptheta' \in \C$ and $\x \in \Univ$,
    \begin{equation} \label{eqn:smooth}
        \norm{\graD{\loss{\x}{\ptheta}} - \graD{\loss{\x}{\ptheta'}}} \leq \smh \norm{\ptheta - \ptheta'}.
    \end{equation}
\end{definition}

\begin{definition}[Strong convexity]
    \label{def:strongly_convex}
    Differentiable function $\loss{\x}{\ptheta}$ is \emph{$\cvx$-strongly convex} if for all $\ptheta, \ptheta' \in~\thetaspace$ and $\x \in \Univ$, 
    \begin{equation} \label{eqn:strongly_convex}
        \loss{\x}{\ptheta'} \geq \loss{\x}{\ptheta} + \graD{\loss{\x}{\ptheta}}^{T} (\ptheta' - \ptheta) + \frac{\cvx}{2} \norm{\ptheta' - \ptheta}^2.
    \end{equation}
\end{definition}

\begin{definition}[Vector field sensitivity]
    \label{def:vector_sensitivity}
    For two vector fields $V, V'$ on $\thetaspace$, we define $\sen{v}$ to be the \mbox{$\ltwo$-sensitivity} between them: 
    \begin{equation} \label{eqn:vector_sensitivity}
        \sen{v} =  \maxu{\ptheta \in \thetaspace} \norm{V(\ptheta) - V'(\ptheta)}.
    \end{equation}
\end{definition}

\begin{definition}[Sensitivity of total gradient]
    \label{def:gradient_sensitivity}
    For a differentiable function $\loss{\x}{\ptheta}$, we define $\sen{\g}$ to be the \mbox{$\ltwo$-sensitivity} of its total gradient $g(\ptheta;\D) = \sum_{\x \in \D}\graD \loss{\x}{\ptheta}$ on neighboring datasets $\D, \D' \in \Domain$: 
    \begin{equation} \label{eqn:gradient_sensitivity}
        \sen{\g} =  \underset{\D\sim\D'}{\max} \ \maxu{\ptheta \in \thetaspace} \norm{g(\ptheta;\D) - g(\ptheta;\D')}.
    \end{equation}
\end{definition}

In Appendix~\ref{ssec:appendix_cal}, we briefly present the basic vector calculus that we require in this paper.

\section{Calculus Refresher}
\label{ssec:appendix_cal}

Given a smooth function $\Loss:\Range \rightarrow \R$, where $\Range \subset \R^d$, its gradient
$\graD{\Loss}: \Domain \rightarrow \R^d$ is the vector of partial derivatives
\begin{equation}
    \label{eqn:grad_dfn}
    \graD{\Loss(\btheta)} = \left(\doh{\Loss(\btheta)}{\btheta_1},\cdots,
    \doh{\Loss(\btheta)}{\btheta_2}\right).
\end{equation}
Its Hessian $\hesS{\Loss}:\Range \rightarrow \R^{d \times d}$ is the matrix of second partial
derivatives
\begin{equation}
    \label{eqn:hess_dfn}
    \hesS{\Loss(\btheta)} = \left(\doh{^2\Loss(\btheta)}{\btheta_i\btheta_j}\right)_{1 \leq i,j \leq d}.
\end{equation}
Its Laplacian $\lapL{\Loss}:\Range \rightarrow \R$ is the trace of its Hessian $\hesS{\Loss}$, i.e.,
\begin{equation}
    \label{eqn:lapl_defn}
    \lapL{\Loss(\btheta)} = \tra{\hesS{\Loss(\btheta)}}.
\end{equation}

Given a smooth vector field $\ve = \left(\ve_1, \cdots, \ve_d\right) :\Range \rightarrow \R^d$, its 
divergence $\divR{\ve}: \Range \rightarrow \R$ is
\begin{equation}
    \label{eqn:divr_defn}
    (\divR{\ve})(\btheta) = \sum_{i=1}^d \doh{\ve_i(\btheta)}{\btheta_i}.
\end{equation}

Some identities that we would rely on:
\begin{enumerate}
    \item Divergence of gradient is the Laplacian, i.e.,
        \begin{equation}
            \label{eqn:divr_lapl_eq}
            \divR{\graD{\Loss}}(\btheta) = \sum_{i=1}^d \doh{^2\Loss(\btheta)}{\btheta_i^2}
            = \lapL{\Loss(\btheta)}.
        \end{equation}
    \item For any function $f: \Range \rightarrow \R$ and a vector field $\ve: \Range \rightarrow
        \R^d$ with sufficiently fast decay to a constant at the border of $\Range$,
        \begin{equation}
            \label{eqn:divr_dotp_eq}
            \int_{\Range} \dotP{\ve(\btheta)}{\graD{f(\btheta)}\dif{\btheta}}
            = - \int_{\Range} f(\btheta)(\divR{\ve})(\btheta)\dif{\btheta}.
        \end{equation}
    \item For any two twice continuously differentiable functions $f, g: \Range \rightarrow \R$, out of which at least for one the gradient decays sufficiently fast at infinity, the following also holds.
        \begin{align}
            \label{eqn:lapl_dotp_eq}
            \int_{\Range} f(\btheta)\lapL{g(\btheta)} \dif{\btheta} 
            &= - \int_{\Range} \dotP{\graD{f(\btheta)}}{\graD{g(\btheta)}} \dif{\btheta} = \int_{\Range} g(\btheta)\lapL{f(\btheta)} \dif{\btheta}.
        \end{align}
        This identity comes from the Green's first identity, which is the higher dimensional equivalent of integration by part.
    \item Based on Young's inequality, for two vector fields $\ve_1, \ve_2: \Range \rightarrow \R^d$,
        and any $a, b \in \R$ such that $ab =1$, the following inequality holds.
        \begin{equation}
            \label{eqn:young_ineq}
            \dotP{\ve_1}{\ve_2} (\btheta) \leq \frac{1}{2a}\norm{\ve_1(\btheta)}^2 
            + \frac{1}{2b}\norm{\ve_2(\btheta)}^2.
        \end{equation}
\end{enumerate}

Wherever it is clear, we would drop $(\btheta)$ for brevity. For example, we would represent
$(\divR{\ve})(\btheta)$ as only $\divR{\ve}$.

\section{Proofs for Section~\ref{sec:privacy_combined}: Privacy analysis of noisy gradient descent}

\subsection{Proofs for Section~\ref{ssec:tracing_diffusion}: Tracing diffusion for Noisy GD}

\begin{replemma}{lem:GDupdate_fokker}
    For coupled tracing diffusion processes \eqr{eqn:soldiscreteSDE} in time $\step\kk < t < \step(\kk+1)$, the equivalent Fokker-Planck equations are
    \begin{equation} \label{eqn:GDupdate_fokker}
    \begin{cases}
    \negthickspace\doh{\rhO_t(\ptheta)}{t} =  \divR{(\rhO_t(\ptheta)V_t(\theta))} + \tnoise^2\lapL{\rhO_t(\ptheta)}\\
    \negthickspace\doh{\rhO_t'(\ptheta)}{t} =  \divR{(\rhO_t'(\ptheta)V'_t(\theta))} + \tnoise^2\lapL{\rhO_t'(\ptheta)},\\
    \end{cases}
    \end{equation}
    where $V_t(\theta) = \Expec{\ptheta_\kk \sim \rhO_{\step\kk|t}}
        {U_2(\ptheta_\kk)|\ptheta}$ and $V'_t(\theta) = \Expec{\ptheta_\kk' \sim \rhO'_{\step\kk|t}}
        {- U_2(\ptheta_\kk)|\ptheta}$ are time-dependent vector fields on $\mathbb{R}^d$, and ${U_2 (\theta)=\frac{1}{2}\left[\nabla \Loss_D(\theta) - \nabla \Loss_{D'}(\theta)\right]}$ is the difference between gradients on neighboring datasets $D$ and $D'$.
\end{replemma}

\begin{proof} We only prove $\negthickspace\doh{\rhO_t(\ptheta)}{t} = \divR{(\rhO_t(\ptheta)V_t(\theta))} + \tnoise^2\lapL{\rhO_t(\ptheta)}$. The proof for the other Fokker-Planck equation is similar.

Recall that conditionals of joint distribution $\rhO_{\step\kk,t}$ is 
    \begin{equation}
        \label{eqn:marg_cond}
        \rhO_{\step\kk,t}(\ptheta_k, \ptheta) = \rhO_{\step\kk}(\ptheta_k) \rhO_{t|\step\kk}(\ptheta|\ptheta_k) = \rhO_{t}(\ptheta) \rhO_{\step\kk|t}(\ptheta_k|\ptheta).
    \end{equation}
    By marginalizing away $\ptheta_k$ in \eqr{eqn:marg_cond}, and taking partial derivative w.r.t. $t$ on both sides, we obtain the following:
    \begin{align*}
        \doh{\rhO_t(\ptheta)}{t} &= \int_{\thetaspace} \doh{\rhO_{t|\step\kk}(\ptheta|\ptheta_k)}{t}\rhO_{\step\kk}(\ptheta_k) \dif{\ptheta_k}  \\
        &=  \int_{\thetaspace} \left(\divR{\left(\rhO_{\step\kk,t}(\ptheta_k,\ptheta)U_2(\ptheta_k)\right)} + \tnoise^2\lapL{\rhO_{\step\kk, t}(\ptheta_k, \ptheta)}\right) \dif{\ptheta_k} \tag{By \eqr{eqn:tracing_fokker} } \\
        &= \divR{\left(\rhO_t(\ptheta)\int_{\thetaspace} \rhO_{\step\kk|t}(\ptheta_k|\ptheta) U_2(\ptheta_k)\dif{\ptheta_k}\right)} + \tnoise^2\lapL{\rhO_t(\ptheta)} \\
        &= \divR{\left(\rhO_t(\ptheta)\Expec{\ptheta_k \sim \rhO_{\step\kk|t}}{U_2(\ptheta_k)|\ptheta}\right)} + \tnoise^2\lapL{\rhO_t(\ptheta)}\\
        &=\divR{\left(\rhO_t(\ptheta)\cdot V_t(\theta)\right)} + \tnoise^2\lapL{\rhO_t(\ptheta)} \tag{where $V_t(\theta)=\Expec{\ptheta_\kk \sim \rhO_{\step\kk|t}}
        {U_2(\ptheta_\kk)|\ptheta}$}
    \end{align*}
\end{proof}

\subsection{Proofs for Section~\ref{ssec:privacy_diffusion}: Privacy erosion in tracing diffusion}

\begin{lemma}[Leibniz integral rule]
    \label{lem:lebnizrule}
    Suppose $f_t: \thetaspace \rightarrow \mathbb \R$ is Lebesgue-integrable for each $t\geq 0$. If for almost all $\ptheta \in \thetaspace$, the derivative $\frac{\dif{f_t}}{\dif{t}}$ exists and there exists an integrable function $g: \thetaspace \rightarrow \R$ such that $\left\vert\frac{\dif{f_t}}{\dif{t}}(\ptheta)\right\vert \leq g(\ptheta)$ for all $t \geq 0$ and almost every $\ptheta \in \thetaspace$, then
    \begin{equation}
        \frac{\dif{}}{\dif{t}}\int_{\thetaspace} f_t(\ptheta) \dif{\ptheta} = \int_{\thetaspace} \frac{\dif{f_t}}{\dif{t}}(\ptheta) \dif{\ptheta}, \quad \text{for all $t \geq 0$}.
    \end{equation}
\end{lemma}

\begin{replemma}{lem:marginalrenyi}[ Rate of R\'enyi divergence ]
    Let $V_t$ and $V_t'$ be two vector fields on $\thetaspace$ with $\maxu{\ptheta \in \thetaspace} \norm{V_t(\ptheta) - V_t'(\ptheta)} \leq \sen{v}$ for all $t\geq 0$. Then, for corresponding coupled diffusions $\{\thet_t\}_{t\geq0}$ and $\{\thet_t'\}_{t\geq0}$ under $V_t$ and $V_t'$ with noise variance $\tnoise^2$, the rate of R\'enyi divergence at any $t\geq0$ is upper bounded by
    \begin{equation}
		\doh{\Ren{\q}{\thet_{t}}{\thet_{t}'}}{t}
		\leq 
	    \frac{1}{\gamma}\frac{\q \sen{v}^2}{4\tnoise^2}
	    -
	    (1-\gamma)\tnoise^2\q \frac{\Gren{\q}{\thet_{t}}{\thet_{t}'}}{\Fren{\q}{\thet_{t}}{\thet_{t}'}}.
    \end{equation}
    where $\gamma>0$ is a tuning parameter that we can fix arbitrarily according to our need.
\end{replemma}

\begin{proof}
    For brevity, let the functions ${R(\q,t)=\Ren{\q}{\rhO_{t}}{\rhO_{t}'}}$, ${E(\q,t)=\Fren{\q}{\rhO_{t}}{\rhO_{t}'}}$, and ${I(\q,t)=\Gren{\q}{\rhO_{t}}{\rhO_{t}'}}$. Under the stated assumptions $\doh{E(\q,t)}{t}$ is bounded as follows.
    \begin{align}
        \label{eqn:marginalrenyi_prfeqn_1}
        \doh{E(\q,t)}{t} &=\doh{}{t}\int_{\thetaspace}\frac{\rhO_{t}^{\q}}{\rhO_{t}'^{\q-1}}\dif{\ptheta} 
    \end{align}  
    By Leibniz integral rule (Lemma~\ref{lem:lebnizrule}), we exchange order of derivative and integration in \eqref{eqn:marginalrenyi_prfeqn_1}. The necessary conditions are satisfied because of the following properties about $p_t$ and $p_t'$:
    \begin{enumerate}
        \item $p_t$ and $p_t'$ have the same support, and their Renyi divergence is well-defined.
        \item The distributions of coupled tracing diffusions $\{\theta_t\}_{\eta k<t<\eta(k+1)}$ and $\{\theta_t'\}_{\eta k<t<\eta(k+1)}$ have  full support and smooth densities $p_t$ and $p_t'$ (due to convolution with Gaussian noise).
        \item The evolutions of probability densities $p_t$ and $p_t'$ with regard to time $t$ satisfy the Fokker-Planck equations~\eqr{eqn:tracing_fokker}.
    \end{enumerate}
    Therefore, we obtain:
    \begin{align*}
        \doh{E(\q,t)}{t} &= \q \int_{\thetaspace} \doh{\rhO_{t}}{t} \left(\frac{\rhO_{t}}{\rhO_{t}'}\right)^{\q-1}\dif{\ptheta} 
        	-(\q-1) \int_{\thetaspace} \doh{\rhO_{t}'}{t} \left(\frac{\rhO_{t}}{\rhO_{t}'}\right)^{\q} \dif{\ptheta} \tag{By Lemma~\ref{lem:lebnizrule}}\\
        	&=\ \q\int_{\thetaspace} \left(\tnoise^2\lapL{\rhO_{t}} + \divR{(\rhO_{t} V_t)}\right)
            \left(\frac{\rhO_{t}}{\rhO_{t}'}\right)^{\q-1}\dif{\ptheta} \\
        &\quad -(\q-1)\int_{\thetaspace} \big(\tnoise^2\lapL{\rhO_{t}'} + \divR{(\rhO_{t}' V_t')}\big) \left(\frac{\rhO_{t}}{\rhO_{t}'}\right)^\q\dif{\ptheta}
            \tag{From \eqr{eqn:GDupdate_fokker}} \\
        &=  \underbrace{
                \tnoise^2 \q\int_{\thetaspace}\left(\frac{\rhO_{t}}{\rhO_{t}'}\right)^{\q-1}\lapL{\rhO_{t}}\dif{\ptheta} \ -\tnoise^2(\q-1)\int_{\thetaspace}\left(\frac{\rhO_{t}}{\rhO_{t}'}\right)^{\q}\lapL{\rhO_{t}'}\dif{\ptheta}
                }_{\eqdef F_1} \\
        &\quad + \underbrace{         
                \q\int_{\thetaspace}\left(\frac{\rhO_{t}}{\rhO_{t}'}\right)^{\q-1} \divR{(\rhO_{t}V_t)}\dif{\ptheta} \ - (\q-1)\int_{\thetaspace}\left(\frac{\rhO_{t}}{\rhO_{t}'}\right)^{\q} \divR{(\rhO_{t}'V_t')}\dif{\ptheta}
            }_{\eqdef F_2}
    \end{align*}
    We simplify $F_1$ as following:
    \begin{align*}
        F_1 &= \tnoise^2(\q-1)\int_{\thetaspace}\dotP{\graD{\left(\frac{\rhO_{t}}{\rhO_{t}'}\right)^\q}}{\graD{\rhO_{t}'}}\dif{\ptheta} -\tnoise^2 \q\int_{\thetaspace}\dotP{\graD{\left(\frac{\rhO_{t}}{\rhO_{t}'}\right)^{\q-1}}}{\graD{\rhO_{t}}}\dif{\ptheta}
                \tag{From \eqr{eqn:lapl_dotp_eq}} \\
            &= \tnoise^2 \q(\q-1)\int_{\thetaspace} \left(\frac{\rhO_{t}}{\rhO_{t}'}\right)^{\q-2} \dotP{\graD{\frac{\rhO_{t}}{\rhO_{t}'}}}
                {\frac{\rhO_{t}}{\rhO_{t}'^2}\graD{\rhO_{t}'} - \frac{\graD{\rhO_{t}}}{\rhO_{t}'}}\rhO_{t}'\dif{\ptheta} \\
            &= -\tnoise^2 \q(\q-1)\Expec{\rhO_{t}'}{\left(\frac{\rhO_{t}}{\rhO_{t}'}\right)^{\q-2} \norm{\graD{\frac{\rhO_{t}}{\rhO_{t}'}}}^2}
                \tag{$\because \graD{\frac{\rhO_{t}}{\rhO_{t}'}} = \frac{\graD{\rhO_{t}}}{\rhO_{t}'} - \frac{\rhO_{t}}{\rhO_{t}'^2}\graD{\rhO_{t}'}$} \\
            &= -\tnoise^2 \q(\q-1) I(\q, t) \tag{From \eqr{eqn:grenyi_divergence}}
    \end{align*}
    We upper bound $F_2$ as following:
    \begin{align*}
        F_2 &= - \q\int_{\thetaspace}\dotP{\graD{\left(\frac{\rhO_{t}}{\rhO_{t}'}\right)^{\q-1}}}{\rhO_{t}V_t}\dif{\ptheta}\ + (\q-1)\int_{\thetaspace}\dotP{\graD{\left(\frac{\rhO_{t}}{\rhO_{t}'}\right)^{\q}}}{\rhO_{t}' V_t'}\dif{\ptheta} 
                \tag{From \eqr{eqn:divr_dotp_eq}} \\
            &= \q(\q-1)\int_{\thetaspace}\left(\frac{\rhO_{t}}{\rhO_{t}'}\right)^{\q-2}\dotP{\graD{\frac{\rhO_{t}}{\rhO_{t}'}}}
                {\frac{\rhO_{t}}{\rhO_{t}'} \left(V_t' - V_t\right)}\rhO_{t}'\dif{\ptheta} \\
            &\leq \gamma \q(\q-1)\tnoise^2\int_{\thetaspace}\left(\frac{\rhO_{t}}{\rhO_{t}'}\right)^{\q-2}\norm{\graD{\frac{\rhO_{t}}{\rhO_{t}'}}}^2\rhO_{t}'\dif{\ptheta}
                \tag{From \eqr{eqn:young_ineq} with $b=2\gamma\tnoise^2$} \\
            &\quad + \frac{\q(\q-1)\sen{v}^2}{4\gamma\tnoise^2}\int_{\thetaspace}\left(\frac{\rhO_{t}}{\rhO_{t}'}\right)^{\q-2}\times\left(\frac{\rhO_{t}}{\rhO_{t}'}\right)^2\rhO_{t}'\dif{\ptheta}
                \tag{$\because \maxu{\ptheta \in \thetaspace} \norm{V_t(\ptheta) - V_t'(\ptheta)} \leq \sen{v}$} \\
            &= \gamma\tnoise^2 \q(\q-1) I(\q,t) + \frac{1}{\gamma}\frac{\q(\q-1)\sen{v}^2}{4\tnoise^2} E(\q,t) \tag{From \eqr{eqn:frenyi_divergence} \& \eqr{eqn:grenyi_divergence}}
    \end{align*}
    Therefore, we get the following bound on the rate of Renyi divergence:
    \begin{align*}
        \doh{R(\q,t)}{t}&=\frac{1}{\q-1}\times \frac{1}{E(\q,t)}\times\doh{E(\q,t)}{t}\\
                                                &\leq -(1-\gamma)\tnoise^2 \q\frac{I(\q,t)}{E(\q,t)} + \frac{1}{\gamma}\frac{\q \sen{v}^2}{4\tnoise^2} 
    \end{align*}
\end{proof}

\paragraphbe{Discussions about the terms in Lemma~\ref{lem:marginalrenyi}} Lemma~\ref{lem:marginalrenyi} bounds the rate of privacy loss with various terms. Generally speaking, the term $\frac{\alpha S_v^2}{4\sigma^2}$ bounds the worst-case privacy loss growth due to noisy gradient update when $S_v=\frac{S_g}{n}$, while the term $\frac{I_{\alpha}(\Theta_t\lVert\Theta_t')}{E_{\alpha}(\Theta_t\lVert\Theta_t')}$ amplifies our bound for the rate of privacy loss, as the Rényi privacy loss accumulates during the process. We offer more explanations as the following.
\begin{enumerate}
    \item $\frac{\alpha S_g^2}{4\sigma^2n^2}$: This is the first term in the right hand side of \eqr{eqn:evolvePDEdiscrete}. It quantifies the worst-case privacy loss due of one noisy gradient update in noisy GD Algorithm \ref{alg:noisygd}. The term $\frac{S_g}{n}$ is the sensitivity of average loss gradient $\Loss_D(\theta)$ over two neighboring datasets $D,D'$. The larger $S_g$ is, the further apart the parameters $\theta$ and $\theta'$ after the gradient descent updates on two neighboring dataset $D,D'$ could be, where $\theta=\theta_0-\eta\nabla \Loss_D(\theta_0)$ and $\theta'=\theta_0-\eta \Loss_{D'}(\theta_0)$. The term $\sigma^2$ is the variance of Gaussian noise. Because additive noise shrink the expected trajectory difference between $\theta$ and $\theta'$ in noisy GD updates, the larger $\sigma^2$ is, the more indistinguishable the distributions of sum of $\theta,\theta'$ and Gaussian noise will be, therefore the smaller the privacy loss (Rényi divergence between end distributions) will be. 
    \item $\frac{I_{\alpha}(\Theta_t\lVert\Theta_t')}{E_{\alpha}(\Theta_t\lVert\Theta_t')}$: This term is the second term in the right hand side of \eqr{eqn:marginalrenyi}, which originates from the derivative of $p_t,p_t'$ with regard to time $t$. To obtain the expression $I_{\alpha}/E_{\alpha}$, we are using the Fokker Planck equation to replace the terms related to $\frac{\partial p_t}{\partial t}, \frac{\partial p_t'}{\partial t}$ with terms determined by the gradient and Laplacian of $p_t,p_t'$ over $\theta$.
    
    The term $I_{\alpha}(\Theta_t\lVert\Theta_t')$ is the \textbf{Rényi information} defined in Definition 2.2., which equals $\mathbb{E}_{\theta\sim p_t'}\left[\left\lVert \nabla \log \frac{p_t(\theta)}{p_t'(\theta)} \right\rVert_2^2 \left(\frac{p_t(\theta)}{p_t'(\theta)}\right)^\alpha\right]$. The term $E_{\alpha}(\Theta_t\lVert\Theta_t')$ is the \textbf{moment of likelihood ratio} defined in Definition 2.1., which equals $\mathbb{E}_{\theta\sim p_t'}\left[\left(\frac{p_t(\theta)}{p_t'(\theta)}\right)^{\alpha}\right]$.  These two terms differ by a \textbf{multiplicative ratio $\left\lVert \nabla \log \frac{p_t(\theta)}{p_t'(\theta)} \right\rVert_2^2$} for their quantities inside expectation. This ratio characterizes the variation of log likelihood ratio function across $\theta$, where $\theta$ is taken from distribution $p_t'$. This is intuitive in the one dimensional case, because $\int_{\theta_1}^{\theta_2}\nabla \log \frac{p_t(\theta)}{p_t'(\theta)} d\theta = \log \frac{p_t(\theta_2)}{p_t'(\theta_2)} - \log \frac{p_t(\theta_1)}{p_t'(\theta_2)}$. Meanwhile since $p_t(\theta),p_t(\theta)'$ are continuous and $\int p_t(\theta)d\theta=\int p_t'(\theta)d\theta=1$, by mean value theorem, there exists $\tilde{\theta}\in\mathbb{R}^d$ such that the log likelihood ratio $\log \frac{p_t(\tilde{\theta})}{p_t'(\tilde{\theta})}$ is zero. Therefore the variation of log likelihood ratio across $\theta$ implicitly increases the largest log likelihood ratio $\max_{\theta\in\mathbb{R}^d}\left[\log(\frac{p_t(\theta)}{p_t(\theta)}) - \log(\frac{p_t(\tilde{\theta})}{p_t(\tilde{\theta})})\right] = \max_{\theta\in\mathbb{R}^d}\left[\log(\frac{p_t(\theta)}{p_t(\theta)})\right]$ across $\theta$ , which reflects the Rényi privacy loss $R_{\alpha}$. 
    
    As a result, intuitively, under some conditions, the larger the Rényi privacy loss $R_{\alpha}$ is, the larger the variation of log likelihood ratio across $\theta$ will be, and therefore the larger the term $\frac{I_{\alpha}(\Theta_t\lVert\Theta_t')}{E_{\alpha}(\Theta_t\lVert\Theta_t')}$ will be. Therefore when the Rényi privacy loss $R_{\alpha}$ is large, the bound for the rate of privacy loss in \eqr{eqn:marginalrenyi} Lemma \ref{lem:marginalrenyi} will also be smaller (under $(1-\gamma)>0$).
    \item $\gamma$ is a tuning constant to balance the privacy growth rate estimated using the above two terms, thus helping us tune the privacy loss accumulation. See the tightness results in Appendix~\ref{sec:appendixtightness} for more details.
\end{enumerate}

\begin{reptheorem}{thm:linearRDP}[Linear R\'enyi divergence bound]
	Let $V_t$ and $V_t'$ be two vector fields on $\thetaspace$, with $\maxu{\ptheta \in \thetaspace} \norm{V_t(\ptheta) - V_t'(\ptheta)} \leq \sen{v}$ for all $t\geq 0$. Then, the diffusion under vector fields $V_t$ and $V_t'$ with noise variance $\sig^2$ for time $T$ has $\q$-R\'enyi divergence of output distributions bounded by $\eps=\frac{\alpha \sen{v}^2 T}{4\sig^2}$.
\end{reptheorem}
\begin{proof}
    Setting $\gamma=1$ in Lemma~\ref{lem:marginalrenyi} gives constant privacy loss rate. Integrating over $t$ suffices.
\end{proof}

\paragraphbe{Controlling R\'enyi privacy loss rate under isoperimetry}

\begin{replemma}{lem:RDinLSI}[~\cite{vempala2019rapid} $\alp$-LSI in terms of Rényi Divergence ]
    Suppose $\thet_t, \thet_t' \in \thetaspace$ are random variables such that probability density ratio between $\thet_t$ and $\thet_t'$ lies in $\Ftheta{\thet_t'}$.
    Then for any $\q \geq 1$, 
    \begin{align} 
        \Ren{\q}{\thet_t}{\thet_t'} 
        + 
        \q(\q - 1) \doh{\Ren{\q}{\thet_t}{\thet_t'}}{\q}  
        \leq 
        \frac{\q^2}{2\alp} \frac{\Gren{\q}{\thet_t}{\thet_t'}}{\Fren{\q}{\thet_t}{\thet_t'}},
    \end{align}
    if and only if $\thet'$ satisfies $\alp$-LSI.
\end{replemma}

\begin{proof} Let $p$ and $p'$ denote the probability density functions of $\thet_t$ and $\thet_t'$ respectively.
    For brevity, let the functions ${R(\q)=\Ren{\q}{\thet_t}{\thet_t'}}$, ${E(\q)=\Fren{\q}{\thet_t}{\thet_t'}}$, and ${I(\q)=\Gren{\q}{\thet_t}{\thet_t'}}$. 
    Let function ${g^2(\ptheta)=\left(\frac{\rhO(\ptheta)}{\rhO'(\ptheta)}\right)^\q}$. Then,
    \begin{align*}
        \Expec{\rhO'}{g^2} = \Expec{\rhO'}{\left(\frac{\rhO}{\rhO'}\right)^{\q}} = \Frenyi{\q}{\rhO}{\rhO'},
        \tag{From \eqr{eqn:frenyi_divergence}}
    \end{align*}
    and,
    \begin{align*}
        \Expec{\rhO'}{g^2 \log g^2} &= \Expec{\rhO'}{\left(\frac{\rhO}{\rhO'}\right)^{\q} 
                                   \log \left(\frac{\rhO}{\rhO'}\right)^{\q}} \\
                                   &= \q \doh{}{\q}\Expec{\rhO'}{\int_{\q}\left(\frac{\rhO}{\rhO'}\right)^{\q} 
                                   \log\left(\frac{\rhO}{\rhO'}\right)\dif{\q}}
                                   \tag{Lebniz's rule} \\
                                   &= \q \doh{}{\q}\Expec{\rhO'}{\left(\frac{\rhO}{\rhO'}\right)^{\q}}
                                   = \q \doh{E(\q)}{\q}.
                                   \tag{From \eqr{eqn:frenyi_divergence}}
    \end{align*}
    Moreover, from \eqr{eqn:grenyi_divergence},
    \begin{align}
        \Expec{\rhO'}{\norm{\graD{g}}^2} = \Expec{\rhO'}{\norm{\graD{
        \left(\frac{\rhO}{\rhO'}\right)^\frac{\q}{2}}}^2}
        = \frac{\q^2}{4} I(\q).
    \end{align}
    On substituting the above equalities in \eqref{eqn:lsi_standard}, we get:
    \begin{align*}
            &\Expec{\rhO'}{g^2 \log g^2} - \Expec{\rhO'}{g^2} \log \Expec{\rhO'}{g^2} \leq \frac{2}{\alp} \Expec{\rhO'}{\norm{\graD{g}}^2} \\
        \iff&\q\doh{E(\q)}{\q} - E(\q) \log E(\q)
            \leq \frac{\q^2}{2\alp} I(\q) \\
        \iff&\q\doh{\log E(\q)}{\q} - \log E(\q) 
            \leq \frac{\q^2}{2\alp} \frac{I(\q)}{E(\q)} \\
        \iff&\q\doh{}{\q}\left((\q-1) R(\q)\right) - (\q - 1)R(\q)
            \leq \frac{\q^2}{2\alp} \frac{I(\q)}{E(\q)}
            \tag{From \eqr{eqn:renyi_divergence}} \\
	\iff& R(\q) + \q(\q-1)\doh{R(\q)}{\q}
            \leq \frac{\q^2}{2\alp} \frac{I(\q)}{E(\q)}
    \end{align*}
\end{proof}

\subsection{Proofs for Section~\ref{ssec:privacy_noisygd}: Privacy guarantee for Noisy GD}
\label{ssec:deferred_privacy_noisygd}

\begin{replemma}{lem:new_delayed_loss_diff}
    Let $\loss{\x}{\ptheta}$ be a loss function on closed convex set $\C$, with a finite total gradient sensitivity $\sen{\g}$. Let $\{\thet_t\}_{t\geq0}$ and $\{\thet_t'\}_{t\geq0}$ be the coupled tracing diffusions for noisy GD on neighboring datasets $\D, \D' \in \Domain$, under loss ${\smaloss{\x}{\ptheta}}$ and noise variance $\tnoise^2$. Then the difference between underlying vector fields $V_t$ and $V_t'$ for coupled tracing diffusions is bounded by
    \begin{equation}
    \label{eqn:new_delayed_loss_diff_appendix}
        \max_{\theta\in\mathbb{R}^d}\Vert V_t(\theta) - V_t'(\theta)\Vert_2 \leq \frac{S_g}{n},
    \end{equation}
    where $V_t(\theta)$ and $V'_t(\theta)$ are time-dependent vector fields on $\mathbb{R}^d$, defined in Lemma~\ref{lem:GDupdate_fokker}.
\end{replemma}

\begin{proof}
    By triangle inequality, for any $\ptheta \in \thetaspace$,
        \begin{align}
            \label{eqn:delay_1}
            \Vert V_t(\theta) - V_t'(\theta)\Vert_2 &\leq \lVert V_t(\theta)\rVert_2 + \lVert V_t'(\theta)\rVert_2 \nonumber \\
            &\leq \frac{1}{2}\Expec{\ptheta_\kk\sim p_{\eta\kk|t}}{\lVert \graD{\Loss_{\D}(\ptheta_\kk)} - \graD{\Loss_{\D'}(\ptheta_\kk)} \rVert_2 \vert\ptheta} \\
            &\quad + \frac{1}{2}\Expec{\ptheta_\kk'\sim p'_{\eta\kk|t}}{\lVert \graD{\Loss_{\D'}(\ptheta_\kk')} - \graD{\Loss_{\D}(\ptheta_\kk')} \rVert_2 \vert\ptheta}. \tag{From Jensen's inequality}
        \end{align}
        By definition of total gradient sensitivity, for any $\theta_k$ and $\theta_k'$, we have 
        \begin{equation*}
            \lVert \graD{\Loss_{\D}(\ptheta_\kk)} - \graD{\Loss_{\D'}(\ptheta_\kk)} \rVert_2\leq \frac{S_g}{\size},\quad \lVert \graD{\Loss_{\D'}(\ptheta_\kk')} - \graD{\Loss_{\D}(\ptheta_\kk')} \rVert_2\leq \frac{S_g}{\size}.
        \end{equation*}
        Therefore, by applying this inequality in equation \eqref{eqn:delay_1} we obtain \eqref{eqn:new_delayed_loss_diff_appendix}.
\end{proof}

\begin{reptheorem}{thm:RDbounddiscrete}
    Let $\{\thet_t\}_{t\geq0}$ and $\{\thet_t'\}_{t\geq0}$ be the tracing diffusion for $\algo_{\text{Noisy-GD}}$ on neighboring datasets $\database$ and $\database'$, under noise variance $\sig^2$ and loss function $\ell(\theta;\x)$. Let $\loss{\x}{\ptheta}$ be a loss function on closed convex set $\C$, with a finite total gradient sensitivity $\sen{\g}$. 
    If for any neighboring datasets $\database$ and $\database'$, the corresponding coupled tracing diffusions $\thet_{t}$ and $\thet_t'$ satisfy $\alp$-LSI throughout ${0\leq t\leq \step\K}$, then $\algo_{\text{Noisy-GD}}$ satisfies $(\q, \eps)$ R\'enyi Differential Privacy for
    \begin{equation} 
        \eps = \frac{\q \sen{\g}^2}{2\alp\sig^4\size^2}(1-e^{-\sig^2\alp\step\K}).
    \end{equation}
\end{reptheorem}

\begin{proof}
    The RDP evolution equation~\eqref{eqn:evolvePDEdiscrete} holds for projected noisy GD during the tracing diffusion in every time piece $\step\kk< t<\step(\kk+1)$. Therefore, for $\eta\kk<t<\step(\kk+1)$, the following differential inequality holds:

    \begin{equation}
        \label{eqn:EvolvePDE_appendix}
        \doh{R(\q,t)}{t}
        \leq
        \frac{1}{\gamma}\frac{\q S_g^2}{4\tnoise^2\size^2}
        - 2(1-\gamma)\sig^2\alp \left[ \frac{R(\q,t)}{\q}     
        + (\q-1)\frac{\partial R(\q,t)}{\partial\q} \right]
    \end{equation}
    
    Let $a_1=2(1-\gamma)\sig^2\alp$, ${a_2=\frac{1}{\gamma}\frac{S_g^2}{4\tnoise^2\size^2}}$, and $y=\log(\q-1)$.
    
    We define the following function $u(t,y)$ based on R\'enyi divergence.
    \begin{align}
        {u(t,y) = \begin{cases}\frac{R(e^y+1,\lim_{t\rightarrow \eta k^+} t)}{e^y+1} - \frac{a_2}{a_1}& if\ t=\eta k\\
            \frac{R(e^y+1,t)}{e^y+1} - \frac{a_2}{a_1}& if\ \eta k<t<\eta(k+1)\\
            \end{cases}}\label{eqn:uty_definition}
    \end{align}
    
    where we denote the limit privacy at start of a step with $R(\q,\lim_{t\rightarrow \eta\kk^{+}}t)=R_{\alpha}(\lim_{t\rightarrow\eta\kk^{+}}\Theta_t\lVert\lim_{t\rightarrow\step\kk^{+}}\Theta_t')$.  Then we can include starting time $t=\eta \kk$ in the time piece for evolution of $u(t,y)$ and re-write \eqr{eqn:EvolvePDE_appendix} as the following:
	\begin{equation}
        \label{eqn:EvolvePDE_appendix_ver_2}
	\frac{\partial u}{\partial t}+a_1 u+a_1\frac{\partial u}{\partial y}\leq 0,\quad \text{when}\ \eta\kk\leq t<\step(\kk+1),
	\end{equation}
	with initial condition 
    \begin{align}
        u(\step\kk,y)=\frac{R(e^y+1,\lim_{t\rightarrow \eta\kk^{+}}t)}{e^y+1}-\frac{a_2}{a_1}
        \label{eqn:uty_initial}
    \end{align}
    
    We introduce auxiliary variables $\tau=t$, and $z=t-\frac{1}{a_1} y$. By defining $v(\tau,z)=u(t,y)$, we get $\frac{\partial v}{\partial \tau}+a_1 v\leq0$ from \eqref{eqn:EvolvePDE_appendix_ver_2}, with initial condition ${v(\step\kk,z) = u(\step\kk,-a_1 (z-\step\kk))}$.  This PDI implies that for every $z$, the rate of decay of $v$ is proportional to its present value. The solution for this PDI is ${v(\tau,z) \leq v(\step\kk,z)e^{-a_1 (\tau-\step\kk)}}$.  By bringing back the original variables, we have
	\begin{equation}
	u(t,y)\leq u(\step\kk,y-a_1 (t-\step\kk))e^{-a_1 (t-\step\kk)},\quad \text{when}\ \eta\kk \leq t<\step(\kk+1).
	\end{equation}
	On undoing the substitution $u(t,y)$ with R\'enyi divergence, via its definition \eqref{eqn:uty_definition} and initial condition \eqref{eqn:uty_initial}, we have that for any $\eta k<t<\eta(k+1)$, the following equation holds.
	\begin{align}
        \frac{R(\alpha,t)}{\alpha} - \frac{a_2}{a_1} = \left(\frac{R((\alpha-1)^{-a_1(t-\eta k)} + 1, \lim_{t_0\rightarrow \eta k^+} t_0)}{(\alpha-1)\cdot e^{-a_1(t-\eta k)} + 1} - \frac{a_2}{a_1}\right)\cdot e^{-a_1(t-\eta k)}.
	\end{align}
    On taking the limit $t\rightarrow\step(\kk+1)^{-}$, we have
    \begin{align}
        \frac{R(\alpha,\lim_{t\rightarrow \eta (k+1)^-} t)}{\alpha} - \frac{a_2}{a_1} = \left(\frac{R((\alpha-1)^{-a_1\eta} + 1,\lim_{t_0\rightarrow \eta k^+} t_0)}{(\alpha-1)\cdot e^{-a_1\eta} + 1} - \frac{a_2}{a_1}\right)\cdot e^{-a_1(t-\eta k)}.
        \label{eqn:recursion}
    \end{align}
    Meanwhile, the tracing diffusion expression~\eqref{eqn:soldiscreteSDE} gives us
    \begin{align}
        \lim_{t\rightarrow\step\kk^{+}}\Theta_t=\phi(\proj{\C}{\lim_{t\rightarrow\step\kk^{-}}\Theta_t}), \quad \text{and} \quad \lim_{t\rightarrow\step\kk^{+}}\Theta_t'=\phi(\proj{\C}{\lim_{t\rightarrow\step\kk^{-}}\Theta_t'}),
    \end{align}
    where $\phi(\theta)=\theta-\step\cdot \frac{1}{2}\left(\Loss_\D(\theta)+\Loss_{\D'}(\theta)\right)$ is a mapping on parameter set $\C\subseteq\mathbb{R}^d$. This mapping is the same for neighboring dataset $\D$ and $\D'$, because its definition only uses the average gradient between neighboring datasets $\D$ and $\D'$. Therefore by post-processing property of R\'enyi divergence, we have that for any $\alpha>1$, the following inequality holds.
    \begin{equation}
        R(\q,\lim_{t\rightarrow\step\kk^{+}}t)\leq R(\q,\lim_{t\rightarrow\step\kk^{-}}t).
        \label{eqn:post_processing}
    \end{equation}
    Combining the above two inequalities~\eqref{eqn:recursion} and \eqref{eqn:post_processing}, we immediately have the following recursive equation:
    \begin{equation}
        \frac{R(\q,\lim_{t\rightarrow\step(\kk+1)^{-}}t)}{\alpha}-\frac{a_2}{a_1}\leq\left(\frac{R((\alpha-1)^{-a_1\eta} + 1,\lim_{t\rightarrow\step\kk^{-}}t)}{(\alpha-1)^{-a_1\eta} + 1}-\frac{a_2}{a_1}\right)e^{-a_1 \step}
    \end{equation}
    Repeating this step for $\kk=0,\cdots,\K-1$, we have
    \begin{equation}
        \label{eqn:rdp_bound_with_limits}
        \frac{R(\q,\lim_{t\rightarrow\step\K^{-}}t)}{\alpha}-\frac{a_2}{a_1}\leq\left(\frac{R(\alpha_0,\lim_{t\rightarrow 0^{-}}t)}{\alpha_0}-\frac{a_2}{a_1}\right)e^{-a_1 \step\K},
    \end{equation}
    
    for some $\alpha_0>1$. Meanwhile, because coupled tracing diffusion have the same start parameter, we have $R(\q_0,\lim_{t\rightarrow0^{-}}t)=0$ for any $\alpha_0$. Moreover, since projection is post-processing mapping, we have $R(\q,\step\K)\leq R(\q,\lim_{t\rightarrow\step\K^{-}}t)$.
    Therefore, taking the value $a_1=2(1-\gamma)\sig^2\alp$, $a_2=\frac{1}{\gamma}\frac{S_g^2}{4\sig^2n^2}$ in \eqref{eqn:rdp_bound_with_limits}, we have
    \begin{equation}
        R(\q,\step\K)\leq \frac{\q S_g^2}{8\gamma(1-\gamma)\alp\sig^4\size^2}(1-e^{-2(1-\gamma)\sig^2\alp\step\K}).
    \end{equation}
    Setting $\gamma=\frac{1}{2}$ suffices to prove the R\'enyi privacy loss bound in the theorem.
\end{proof}

\paragraphbe{Isoperimetry constants for noisy GD}

To prove that LSI holds for the tracing diffusion for noisy GD, we first note that the diffusion process~\eqr{eqn:soldiscreteSDE} can be written as composition of Lipschitz mapping and Gaussian noise for any $\eta k< t< \eta (k+1)$. Meanwhile, the projection at the end of a step is $1$-Lipstchitz mapping. Then, we rely on the following two lemmas that show Lipschitz transformation and Gaussian perturbation of a probability distribution preserve its LSI property.

\begin{lemma}[LSI under Lipschitz transformation~\cite{ledoux2001concentration}]
    \label{lem:lip_lsi}
    Suppose a probability distribution $\rhO$ on $\thetaspace$ satisfies LSI with constant~$\alp > 0$.  Let $T:\thetaspace \rightarrow \thetaspace$ be a differentiable and $\lip$-Lipschitz transformation. The push-forward distribution~$T_{\#\rhO}$, representing $T(\thet)$ when $\thet \sim \rhO$, satisfies LSI with constant~$\frac{\alp}{\lip^2}$.
\end{lemma}

\begin{lemma}[LSI under Gaussian convolution~\cite{ledoux2001concentration}]
    \label{lem:gaus_conv_lsi}
    Suppose a probability distribution $\rhO$ on $\thetaspace$ satisfies LSI with constant~${\alp > 0}$.  For $t>0$, the probability distribution $\rhO * \Gaus{0}{2t\Id}$ satisfies LSI with constant~$\left(\frac{1}{\alp} + 2t\right)^{-1}$. A special case of this is that $\Gaus{0}{2t\Id}$ satisfies LSI with constant~$\frac{1}{2t}$.
\end{lemma}

\begin{replemma}{lem:iso_convex}
    If loss function $\loss{\x}{\ptheta}$ is $\cvx$-strongly convex and $\smh$-smooth over a closed convex set $\C$, the step-size is ${\step < \frac{1}{\smh}}$, and initial distribution is $\thet_{0} \sim \proj{\C}{\Gaus{0}{\frac{2\tnoise^2}{\cvx}\Id}}$, then the coupled tracing diffusion processes $\{\thet_t\}_{t\geq 0}$ and $\{\thet_t'\}_{t\geq 0}$ for noisy GD on any neighboring datasets $\D$ and $\D'$ satisfy $\alp$-LSI for any $t\geq0$ with $\alp=\frac{\lam}{2\sig^2}$.
 \end{replemma}
 
 \begin{proof}
    We only prove $\alp$-LSI for the tracing diffusion process $\{\thet_t\}_{t\geq 0}$ on dataset $\D$. The proof for $\{\thet_t'\}_{t\geq 0}$ is similar.
 
    For any $\D\in \Domain$, and any $0<s<\step$, recall that the update step in tracing diffusion~\eqref{eqn:soldiscreteSDE} equals the following random mapping: 
    \begin{equation}
        \label{eqn:pushover}
        \Theta_{\step\kk+s}=\begin{cases}T_s(\Theta_{\step\kk})+\sqrt{2s\sig^2}\gaussnoise,&\text{if}\ 0\leq s<\eta\\\proj{\C}{ T_s(\Theta_{\step\kk})+\sqrt{2s\sig^2}\gaussnoise},&\text{if}\ s=\eta\end{cases}
    \end{equation}
    where the mapping ${T_s(\ptheta)=\ptheta- \eta\cdot \frac{1}{2}\left(\nabla\Loss_\D(\theta)+\nabla\Loss_{\D'}(\theta)\right)} -s\cdot \frac{1}{2}\left(\nabla\Loss_\D(\theta)-\nabla\Loss_{\D'}(\theta)\right) $. We first show that $T_s(\theta)$ is $(1-\eta\cvx)$-Lipschitz. For any $w, v \in \C$, we have
     \begin{align*}
         T_s(w)-T_s(v) &= w-v-\frac{\eta+s}{2}[\nabla\Loss_\D(w)-\nabla\Loss_\D(v)] - \frac{\eta-s}{2}[\nabla\Loss_{\D'}(w)-\nabla\Loss_{\D'}(v)]\\
         &=w-v- \left[\frac{\eta+s}{2}\hesS{\Loss_\D(z)}+\frac{\eta-s}{2}\hesS{\Loss_{\D'}(z')}\right](w-v) \tag{for some $z,z' \in \C$ by the mid-value theorems}\\
         &=\Big(I-\left[\frac{\eta+s}{2}\hesS{\Loss_\D(z)}+\frac{\eta-s}{2}\hesS{\Loss_{\D'}(z')}\right]\Big)(w-v)
     \end{align*}  
     By $\lam$-strong convexity and $\smh$-smoothness of loss function $\ell(\theta;\x)$ on $\C$, we prove that $\hesS{\Loss_\D(z)}$ and $\hesS{\Loss_{\D'}(z')}$ both have eigenvalues in the range $[\cvx,\smh]$. Since $s<\step<\frac{1}{\smh}$, all eigenvalues of $I- \left[\frac{\eta+s}{2}\hesS{\Loss_\D(z)}+\frac{\eta-s}{2}\hesS{\Loss_{\D'}(z')}\right]$ is in $(0,1-\eta\cvx]$. So, $T_s$ is $(1-\eta\cvx)$-Lipshitz. 
     
     Now, using induction we prove $\rhO_t$ statisfies $\alp$-LSI for $\alp=\frac{\cvx}{2\tnoise^2}$ for any $t \geq 0$. 
     
     {\bf Base step:} Being a projection of Gaussian with variance $\frac{\cvx}{2\tnoise^2}$ in every dimension, $\thet_0$ satisfies $\alp$-LSI with the given constant by Lemma~\ref{lem:lip_lsi} (because projection is $1$-Lipschitz) and Lemma \ref{lem:gaus_conv_lsi}.    
 
     {\bf Induction step:} Suppose $\thet_{\step\kk}$ satisfies $\alp$-LSI with the above constant for some $k \in \N$. Distribution $\thet_{t}$ for ${\step \kk < t < \step (\kk+1)}$ is same as $T_s$ pushover distribution plus gaussian noise distribution, i.e.  ${\thet_t = \thet_{\step \kk}}_{\# T_s} * \Gaus{0}{2s\tnoise^2\Id}$ for $s = t - \step \kk$. By using Lemma \ref{lem:lip_lsi} and \ref{lem:gaus_conv_lsi}, we get
     $\left( \frac{\alp}{(1 - \eta\cvx)^2 + 2s\tnoise^2\alp} \right)$-LSI for $\thet_t$.
     Since $s < \step < \frac{1}{\cvx}$, we have
     \begin{equation*}
         (1-s\cvx)^2 + 2s\tnoise^2\alp < 1 - s\cvx + 2s\tnoise^2\alp = 1.
     \end{equation*}
     Hence, for $\eta\kk < t<\eta(\kk+1)$, $\thet_{t}$ satisfies $\alp'$-LSI with constant $\alp' > \alp$, which means it also satisfies $\alp$-LSI by definition.
     
     By \eqr{eqn:pushover}, $\thet_{\eta(k+1)}$ undergoes an additional projection $\proj{\C}{\cdot}$. Since projection is a 
     $1$-Lipschitz map, by Lemma \ref{lem:lip_lsi}, it preserves $\alp$-LSI. So distribution $\thet_{\eta(k+1)}$ also satisfies $\alp$-LSI.
 \end{proof}

\section{Proofs and discussions for Section~\ref{ssec:tightness}: Tightness analysis}

\label{sec:appendixtightness}

\begin{reptheorem}{thm:tightness}
    There exist two neighboring datasets $\D, \D' \in \Domain$, a start distribution $\rhO_0$, and a smooth loss function $\loss{\x}{\ptheta}$ whose total gradient $g(\ptheta;\D)$ has finite sensitivity $\sen{\g}$ on unconstrained convex set $\C=\mathbb{R}^d$, such that for any step-size $\step<1$, noise variance $\tnoise^2>0$, and $\K\in\mathbb{N}$, the R\'enyi privacy loss of $\algo_{\text{Noisy-GD}}$ on $\D, \D'$ is lower-bounded by
    \begin{equation}
        \Ren{\q}{\thet_{\step\K}}{\thet_{\step\K}'} \geq \frac{\q S_g^2}{4\sig^2 n^2} \left( 1 - e^{-\step\K} \right).
    \end{equation}
\end{reptheorem}

\begin{proof}

    We give lower bounds for the R\'enyi DP guarantee of noisy gradient descent algorithm for minimizing any smooth loss function $\loss{\x}{\ptheta}$ with finite total sensitivity $S_g$. We consider the following $L_2$-norm squared loss function with bounded data universe.

    \begin{equation} \label{eqn:lossl2normsquare}
        \loss{\x}{\ptheta} = \frac{1}{2} \norm{\ptheta - \x}^2, \text{where} \ \ptheta\in\thetaspace, \x \in \thetaspace\text{ and } \norm{\x} \leq \frac{S_g}{2}.
    \end{equation}
    
    For any dataset $D=\{\x_1,\cdots,\x_n\}$ of size $n$, and any $\ptheta\in \thetaspace$, the loss is
    \begin{equation*}
        \Loss_\D(\ptheta)=\frac{1}{n}\sum_{i=1}^n\frac{1}{2}\norm{\ptheta-\x_i}^2.
    \end{equation*}
    
    It is easy to verify that $\Loss_\D(\ptheta)$ is 1-smooth. The total gradient of $\D$ is
    \begin{align*}
        \g(\ptheta;\D) = \sum_{\x \in \D} \graD{\loss{\x}{\ptheta}} = \size\ptheta - \sum_{\x \in \D} \x,
    \end{align*} 
    with a finite sensitivity $\sen{\g}$.
    
    We construct the two neighboring datasets $\D, \D' \in \Domain$ such that ${\D = (\x_1, 0^d, \cdots, 0^d)}$ and ${\D' = (\x_1', 0^\dime, \cdots, 0^d)}$, where $\x_1, \x_1' \in \Univ$ are two records that are $S_g$ distance apart (i.e. $\norm{\x_1 - \x_1'} = S_g$).
    
    Under dataset $\D$, we can express the random variable $\thet_{\step\K}$ at the $\K$'th iteration of noisy GD using the following recursion with starting parameter $\Theta_0 = 0^\dime$.
    \begin{align*}
        \thet_{\step\K} =& (1 - \step) \thet_{\step (\K-1)} +\step \frac{\x_1}{n} + \sqrt{2 \step \tnoise^2} \cdot \Z_{\K - 1} \\
        =& (1-\step)^K\thet_0+\step\sum_{i=0}^{\K-1}(1-\step)^i\frac{\x_1}{n} + \sqrt{2\step\tnoise^2}\sum_{i=0}^{\K-1}(1-\step)^{\K-1-i} \Z_{i} \\
        %
        %
        =& \frac{\step\x_1}{\size}\sum_{i=0}^{\K-1}(1-\step)^{i}
        + \sqrt{2\step\tnoise^2\sum_{i=0}^{\K-1}(1 - \step)^{2i}} \cdot \Z \tag{where $\Z_i, \Z \sim \Gaus{0}{\Id}$}
    \end{align*}
    
    A similar recursion can be used for $\thet_\K'$ in Noisy GD under dataset $D'$.  Both $\thet_\K$ and $\thet_\K'$ are Gaussian random variables with variance $2\step\tnoise^2\sum_{i=0}^{\K-1}(1 - \step)^{2i}$ in each dimension.  Thus, we can calculate their exact divergence.
    \begin{align*}
    \Renyi{\q}{\thet_{\step\K}}{\thet_{\step\K}'} &= \frac{\q\cdot \norm{\step(\x_1-\x_1')\sum_{i=0}^{\K-1}(1-\step)^{i}}^2}{2\cdot 2\step\sig^2n^2\sum_{i=0}^{\K-1}(1-\step)^{2i}} \\
        &= \frac{\q \step^2S_g^2}{4\step\sig^2n^2} \cdot \frac{\left(1-(1-\step)^\K\right)^2/\step^2}{(1-(1-\step)^{2\K})/\left(\step(2-\step)\right)} \\
        &= \frac{\q S_g^2}{4\sig^2n^2} \cdot \frac{2-\step}{1+(1-\step)^{\K}} \left(1-(1-\step)^\K\right)\\
        &\geq \frac{\q S_g^2}{4\sig^2n^2} (1-e^{-\step\K})
    \end{align*}
    This inequality concludes the proof.
\end{proof}

\begin{repcorollary}{cor:tightness}
Given $\ell_2$-norm squared loss function ${\ell(\theta;\x)=\frac{1}{2}\norm{\theta-\x}^2}$ on unconstrained convex set $\C=\mathbb{R}^d$ and bounded data domain with range $S_g$, and initial parameter $\theta_0=0^\dime$, for any two neighboring datasets $\D, \D' \in \Domain$, step-size $\step$, noise variance $\tnoise^2$, and $\K \in \N$, the R\'enyi privacy loss of $\algo_{\text{Noisy-GD}}$ on $\D, \D'$ is upper-bounded by
\begin{equation}
    \Ren{\q}{\thet_{\step\K}}{\thet_{\step\K}'} \leq \frac{\q S_g^2 }{(2-\step)\sig^2n^2}(1-e^{-\frac{2-\step}{2}\step\K}).
\end{equation}
\end{repcorollary}
\begin{proof}
    To use Theorem~\ref{thm:RDbounddiscrete}, we still need to verify $\alp$-LSI for the tracing diffusion on $\ell_2$-norm squared loss. 

    We use the explicit expression for tracing diffusion proved in Theorem~\ref{thm:tightness} to prove $\alp$-LSI. We utilize the fact that $\thet_{\step\K}$, the tracing diffusion for $L_2$-norm squared loss at discrete update time $\step K$, is Gaussian with bounded variance ${2\step\tnoise^2\sum_{i=0}^{\K-1}(1 - \step)^{2i}\leq \frac{2\sig^2}{2-\eta}}$ in each dimension.  Therefore, based on Lemma~\ref{lem:gaus_conv_lsi}, which shows the LSI properties of Gaussian distributions, $\thet_{\K\step}$ satisfies $\alp$-LSI with $\alp=\frac{2-\step}{2\sig^2}$.  Similarly, by computing the explicit expression for tracing diffusion at time $\step \kk < t < \step (\kk+1)$, one can verify $\thet_t$ satisfies $\alp$-LSI.
    
    Now, we can directly use Theorem~\ref{thm:RDbounddiscrete} to derive an upper-bound for RDP for Noisy GD under $L_2$-squared norm loss.
\end{proof}

\paragraphbe{Discussion about tightness results}

\pgfplotsset{
	/pgf/declare function={
		lower_bound(\k,\a,\e,\r,\s,\n) = \a*\r^2/(2*2*\s^2*\n^2)*(1-exp(-\k*\e));
	}
}

\pgfplotsset{
	/pgf/declare function={
		rdp_bound(\k,\a,\e,\r,\s,\n) = \a*\r^2/((2-\e)*\s^2*\n^2)*(1-exp(-(2-\e)/2*\k*\e));
	}
}

\pgfplotsset{
	/pgf/declare function={
		comp_bound(\k,\a,\e,\r,\s,\n) = \a*\r^2/(2*2*\s^2*\n^2)*\e*\k;
	}
}

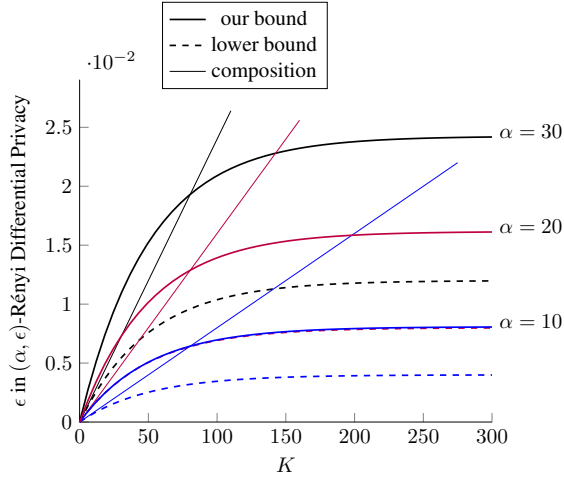
\begin{figure}[h]
    \centering 
    \begin{tikzpicture}[scale=0.8]
        \begin{axis}[
        no markers,
        samples=50,
        xmin=0,
        ymin=0,
        xlabel=$K$,
        ylabel={$\epsilon$ in $(\q, \epsilon)$-R\'enyi Differential Privacy},
        xmax=300,clip=false,axis y line*=left,axis x line*=bottom,legend style={at={(0.2,1.1)},anchor=west}]
            \addplot[thick,black,domain=0:300] {rdp_bound(x,30,0.02,4,0.02,5000)};
            \addlegendentry{our bound}
            \addplot[thick,black,dashed,domain=0:300] {lower_bound(x,30,0.02,4,0.02,5000)};
            \addlegendentry{lower bound}
            \addplot[black,domain=0:110] {comp_bound(x,30,0.02,4,0.02,5000)};
            \addlegendentry{composition}
            \addplot[blue,domain=0:275] {comp_bound(x,10,0.02,4,0.02,5000)};
            \addplot[purple,domain=0:160] {comp_bound(x,20,0.02,4,0.02,5000)};
            \addplot[thick,blue,dashed,domain=0:300] {lower_bound(x,10,0.02,4,0.02,5000)};
            \addplot[thick,purple,dashed,domain=0:300] {lower_bound(x,20,0.02,4,0.02,5000)};
            \addplot[thick,blue,domain=0:300] {rdp_bound(x,10,0.02,4,0.02,5000)};
            \addplot[thick,purple,domain=0:300] {rdp_bound(x,20,0.02,4,0.02,5000)};
            \node[anchor=west] at (axis cs: 300,{rdp_bound(300,10,0.02,4,0.02,5000)+0.0005}) {$\q=10$};
            \node[anchor=west] at (axis cs: 300,{rdp_bound(300,20,0.02,4,0.02,5000)+0.0005}) {$\q=20$};
            \node[anchor=west] at (axis cs: 300,{rdp_bound(300,30,0.02,4,0.02,5000)+0.0005}) {$\q=30$};
        \end{axis}
    \end{tikzpicture}
    \caption{Tightness analysis of our RDP guarantee for the noisy GD algorithm. We show the changes of $\q$-RDP guarantee computed using Corollary~\ref{cor:tightness}, over $K$ iterations (number of full passes over the dataset) versus the lower-bounds (dashed lines) which are computed using Theorem~\ref{thm:tightness}.  The loss function is the $\ell_2$-norm squared function~\eqref{eqn:lossl2normsquare}, noise standard deviation is $\sig=0.02$, the step size is $\eta=0.02$, the size of the dataset is $\size=5000$, and the finite $\ell_2$-sensitivity for total gradient is $S_g=4$. The expression for computing the privacy loss in Baseline composition-based analysis (derived by moment accountant~\cite{abadi2016deep} with details in Appendix~\ref{sec:appendixtightness}) is: ${\eps=\frac{\q S_g^2}{4n^2\sig^2}\cdot\step\K}$} \label{fig:tightness}
\end{figure}

Figure~\ref{fig:tightness} shows the gap between this lower bound and our RDP guarantee derived by Corollary~\ref{cor:tightness}, under small step-size $\eta=0.02$. The upper bound is roughly two times larger than the lower bound, which shows tightness of our privacy guarantee up to a rough constant of two. As comparison, we compute and plot the composition-based bound, which grows as fast as the lower bound in early iterations, but linearly grows above the lower bound, and our RDP guarantee, as $\K$ increases to ${\Omega(\frac{1}{\eta})\approx100\ll \size=5000}$. Moreover, the larger the RDP order $\q$ is, the smaller the required number of iterations $\K$ is for our RDP guarantee to be superior to the composition-based privacy bound.

\paragraphbe{Gap between our upper bound and lower bound}  There is a gap between the exponent and constant of our privacy upper bound Corollary~\ref{cor:tightness} and the lower bound Theorem~\ref{thm:tightness}. We analyze the gap as follows.
\begin{enumerate}
    \item \textbf{The gap in exponent:} There is a $\frac{2-\eta}{2}$ multiplicative gap between the exponent of our privacy upper bound and the lower bound. In hindsight, this is because discretized noisy GD converges to a biased stationary distribution. Therefore, our LSI constant bound $c=\frac{2-\eta}{2\sigma^2}$ depends on the discretization bias caused by step-size $\eta$, thus causing the exponent gap in our privacy bound.
    \item \textbf{The gap in constant:} Our upper bound is larger than the lower bound by roughly a multiplicative constant of two. This is due to the \textbf{balancing ratio $\gamma>0$} in Lemma~\ref{lem:marginalrenyi} for bounding the rate of privacy loss. 
    \begin{enumerate}
        \item \textbf{At the start of Noisy GD:} setting $\gamma=1$ in \eqr{eqn:marginalrenyi} results in a smaller privacy loss rate bound. This is because, at the start of noisy GD, the accumulated privacy loss $R_{\alpha}(\Theta_t\lVert\Theta_t')$ is small, thus leading to a small second term $I_{\alpha}/E_{\alpha}$ in \eqr{eqn:marginalrenyi}, by Lemma ~\ref{lem:RDinLSI}. Setting $\gamma=1$ reduces the coefficient $\frac{1}{\gamma}$ for the dominating first term of \eqr{eqn:marginalrenyi}, at a small cost of increasing the coefficient for the smaller second term $I_{\alpha}/E_{\alpha}$. This facilitates a smaller privacy loss rate bound, and is reflected in the similar growth of composition bound (equivalent to setting $\gamma=1$) and our lower bound in Figure ~\ref{fig:tightness}.
        \item \textbf{As Noisy GD converges: }setting $\gamma\rightarrow 0$ in \eqr{eqn:marginalrenyi} results in a smaller privacy loss rate bound. This is because, at convergence, the accumulated privacy loss $R_{\alpha}(\Theta_t\lVert\Theta_t')$ is larger, thus leading to more significant second term $I/E$ in \eqr{eqn:marginalrenyi}. Setting $\gamma\rightarrow 0$ in \eqr{eqn:marginalrenyi} reduces the coefficient $-(1-\gamma)$ for the dominating second term $I/E$, thus facilitate a smaller bound for the privacy loss rate.
    \end{enumerate}
    \item In our proof for Theorem ~\ref{thm:RDbounddiscrete}, we set $\gamma=\frac{1}{2}$ to \textbf{balance} privacy loss rate estimates at the start and convergence of noisy GD, thus obtaining the smallest privacy bound at convergence, as shown in the proof.
\end{enumerate}
\paragraphbe{Derivation for Baseline composition-based privacy bound}
\citet{abadi2016deep} introduce the moments accountant $\alpha(\lambda)$ for noisy SGD in Eq (2) of their paper, which effectively tracks the scaled Renyi divergence between processes. Therefore in Figure~\ref{fig_ourbound}, we plot moment accountant bound in \citet{abadi2016deep} as baseline composition privacy analysis. 
\begin{enumerate}
    \item We first use \textbf{moments bound on the Gaussian mechanism} (following Lemma 3 in \citet{abadi2016deep} ) to bound the log moment $\alpha_{\mathcal{M}}(\lambda)$ of data-sensitive computation one update $M: \mathcal{M}(D) = \frac{\eta}{n} \sum_{x_i\in D} \nabla \ell(\theta;x_i) + \mathcal{N}( 0 , 2\eta\sigma^2 \mathbb{I}_d )$ in our Algorithm ~\ref{alg:noisygd}. 
    
    By Eq (2) in \citet{abadi2016deep}, and that $M(D),M(D')$ are Gaussian distributions (with variance $2\eta\sigma^2$ in every dimension and means at most $\frac{\eta}{n} S_g$ apart in $\ell_2$ norm), we bound $\alpha_{\mathcal{M}}(\lambda) \leq \frac{\lambda (\lambda+1) \eta S_g^2}{4n^2\sigma^2}$.
    \item We then \textbf{compose log moment bound for $K$ iterations} by Theorem 2 [Composibility] of log moement bound in \citet{abadi2016deep}., and we obtain $\alpha(\lambda)\leq K \cdot \alpha_{\mathcal{M}}(\lambda)=\frac{K\lambda(\lambda+1)\eta S_g^2}{4n^2\sigma^2}$. 
    \item Finally by definition of log moment (Eq (2) of \citet{abadi2016deep}) and Renyi divergence (\eqref{eqn:renyi_divergence_short_prelim} in our paper), we take $\lambda \leftarrow \alpha-1$ and $R_{\alpha}(\Theta_K\lVert\Theta_K) \leftarrow \frac{\alpha(\lambda)}{\lambda}$, and obtain the \textbf{baseline composition privacy bound} $\epsilon = \frac{\alpha S_g^2}{4n^2\sigma^2}\cdot \eta K$ from the log moment bound. We use this expression in Figure~\ref{fig_ourbound} and \ref{fig:tightness}.
\end{enumerate}

\section{Proofs for Section \ref{ssec:utility}: Utility analysis}
\label{sec:projectionproof}
%

\begin{reptheorem}{thm:projectedexcessrisk}
	For Lipschitz smooth strongly convex loss function $\ell(\theta;\x)$ on a bounded closed convex set $\mathcal{C}\subseteq\thetaspace$, and dataset $\D\in\Domain$ of size $n$, if the step-size $\eta=\frac{\lambda}{2\beta^2}$ and the initial parameter $\theta_0\sim\proj{\C}{\mathcal{N}(0,\frac{2\sig^2}{\lambda}\mathbb{I}_d)}$, then the noisy GD Algorithm~\ref{alg:noisygd} is $(\alpha,\eps')$-R\'enyi differentially private, where $\alpha>1$ and $\eps'>0$, and satisfies
	\begin{equation}
		\mathbb{E}[\Loss_D(\theta_{K^*})-\Loss_D(\theta^*)]=O(\frac{\alpha\beta d L^2}{\eps'\lam^2\size^2}),
	\end{equation}
	by setting noise variance $\sig^2=\frac{4\alpha \Lip^2}{\lambda\eps' n^2}$, and number of updates $K^*=\frac{2\beta^2}{\lam^2}\log(\frac{n^2\eps'}{\alpha d})$.
	
	Equivalently, for $\eps\leq 2\log(1/\delta)$ and $\delta>0$, Algorithm~\ref{alg:noisygd} is $(\eps,\delta)$-differentially private, and satisfies
	\begin{equation}
		\mathbb{E}[\Loss_D(\theta_{K^*})-\Loss_D(\theta^*)]=O(\frac{\beta d L^2\log(1/\delta)}{\epsilon^2\lam^2\size^2}),
	\end{equation}
	by setting noise variance $\sig^2=\frac{8\Lip^2 (\eps+2\log(1/\del))}{\lambda\eps^2 n^2}$, and number of updates $K^*=\frac{2\beta^2}{\lam^2}\log(\frac{n^2\eps^2}{4\log(1/\delta) d})$.
\end{reptheorem}

\begin{proof}
	From Lemma~\ref{lem:pointwise_conv}, we have
	\begin{equation}
		\label{eqn:utility_bound1}
		\mathbb{E}[\Loss_D(\theta_{K})-\Loss_D(\theta^*)]\leq  \frac{2\beta L^2}{\lambda^2}e^{-\lambda\eta K} +\frac{2\beta d\sigma^2}{\lambda}.
	\end{equation}
	Since $\eta=\frac{\lam}{2\beta^2}\leq \frac{1}{\beta}$, by Corollary~\ref{cor:main_theorem}, the noisy GD with $K$ iterations will be $(\alpha,\eps')$-RDP as long as $\sig^2\geq\frac{4\alpha \Lip^2}{\lambda\eps' n^2}(1-e^{-\lambda\eta K/2})$. Therefore, if we set $\sig^2= \frac{4\alpha\Lip^2}{\lambda \eps' n^2}$, noisy GD is $(\alpha,\eps')$-RDP for any $K$. On substituting this noise variance in \eqref{eqn:utility_bound1}, we get
	\begin{equation}
		\mathbb{E}[\Loss_D(\theta_{K})-\Loss_D(\theta^*)]\leq  \frac{2\beta L^2}{\lambda^2}e^{-\lambda\eta K} +\frac{8\alpha\Lip^2\beta d}{\lambda^2\eps' n^2}.
	\end{equation}
	By setting $K^*=\frac{1}{\lam\eta}\log(\frac{\eps' n^2}{\alpha d})=\frac{2\beta^2}{\lam^2}\log(\frac{\eps' n^2}{\alpha d})$, we can control the empirical risk to be
	\begin{equation}
		\label{eqn:utility_bound2}
		\mathbb{E}[\Loss_D(\theta_{K^*})-\Loss_D(\theta^*)]\leq  \frac{10\alpha\Lip^2\beta d}{\lambda^2\eps' n^2}.
	\end{equation}
	Now, we convert the optimal excess risk guarantee under an $(\q, \eps')$ RDP constraint to an optimal excess risk guarantee under $(\eps, \del)$ DP constraint. Let $\eps> 0$ and  $0<\del<1$ be two constants such that $\eps\leq 2\log(1/\del)$. As per DP transition Theorem~\ref{thm:RDP_to_DP}, $(\q, \eps')$-RDP implies $(\eps,\del)$-DP for $\alpha=1+\frac{2}{\eps}\log(1/\delta)$ and $\eps'=\frac{\eps}{2}$. By using this conversion, we bound \eqref{eqn:utility_bound2} in terms of DP parameters as 
	\begin{align*}
		\mathbb{E}[\Loss_D(\theta_{K^*})-\Loss_D(\theta^*)]&\leq  \frac{10\Lip^2\beta d}{\lambda^2n^2}\frac{\alpha}{\eps'}\\
		&=\frac{10\Lip^2\beta d}{\lambda^2n^2}\frac{1+\frac{2}{\eps}\log(1/\del)}{\frac{\eps}{2}}\\
		\because\eps\leq 2\log(1/\del)\quad &\leq \frac{10\Lip^2\beta d}{\lam^2n^2}\frac{8\log(1/\del)}{\eps^2}.\\
	\end{align*}
	The amount of noise needed in terms of DP parameters is
	\begin{align*}
		\sig^2&=\frac{4\Lip^2}{\lam n^2}\frac{\q}{\eps'}\\
		&=\frac{4\Lip^2}{\lam n^2}\cdot\frac{1+\frac{2}{\eps}\log(1/\del)}{\frac{\eps}{2}}
	\end{align*}
	The optimal number of updates $K^*$ in terms of DP parameters is bounded as
	\begin{align*}
		\K^*&=\frac{2\beta^2}{\lam^2}\log(\frac{n^2}{d}\cdot\frac{\eps'}{\alpha})\\
		&=\frac{2\beta^2}{\lam^2}\log(\frac{n^2}{d}\cdot\frac{\frac{\eps}{2}}{1+\frac{2}{\eps}\log(1/\del)})\\
		&\leq\frac{2\beta^2}{\lam^2}\log(\frac{n^2}{d}\cdot\frac{\eps^2}{4\log(1/\del)}).
	\end{align*}
\end{proof}

\begin{replemma}{lem:pointwise_conv}
	For $\Lip$-Lipschitz, $\lambda$-strongly convex and $\beta$-smooth loss function $\ell(\theta;\x)$ over a closed convex set $\C\subseteq\mathbb{R}^d$, step-size $\eta\leq \frac{\lambda}{2\beta^2}$, and start parameter $\theta_0\sim\proj{\C}{\mathcal{N}(0,\frac{2\sig^2}{\lam}\Id)}$, the excess empirical risk of Algorithm~\ref{alg:noisygd} is bounded by
	\begin{equation}
		\mathbb{E}[\Loss_{\database}(\theta_{K}) - \Loss_{\database}(\theta^*)] \leq  \frac{2\beta L^2}{\lambda^2}e^{-\lambda\eta K} +\frac{2\beta d\sigma^2}{\lambda},
	\end{equation}
	where $\theta^*$ is the minimizer of $\Loss_\D({\theta})$ in the relative interior of convex set $\mathcal{C}$, and $d$ is the dimension of parameter.
\end{replemma}

\begin{proof}
	By the noisy GD update equation we have 
	\begin{equation}
		\theta_{k+1}=\proj{\C}{\theta_k-\eta\nabla\Loss_\D(\theta_k)+\sqrt{2\eta\sig^2}\mathcal{N}(0,\mathbb{I}_d)}.
	\end{equation}
	From the definition of projection $\proj{\C}{\cdot}$, we have:
	\begin{align*}
		\proj{\C}{\theta^*-\eta\nabla\Loss_{\D}(\theta^*)}&=\arg\min_{\theta\in\C}\lVert\theta-\theta^*+\eta\nabla\Loss_\D(\theta^*)\rVert_2^2\\
		&=\arg\min_{\theta\in\C}\lVert\theta-\theta^*\lVert_2^2
		+2\eta\langle\theta-\theta^*,\nabla\Loss_\D(\theta^*)\rangle
		+\eta^2\lVert\nabla\Loss_\D(\theta^*)\rVert_2^2
		\tag{by optimality of $\theta^*$ in $\mathcal{C}$} \\
		&=\arg\min_{\theta\in\C}\lVert\theta-\theta^*\lVert_2^2
		+\eta^2\lVert\nabla\Loss_\D(\theta^*)\rVert_2^2\\
		&=\theta^*
	\end{align*}

	Therefore, by combining the above two, and from contractivity of projection $\proj{\C}{\cdot}$~\cite[Proposition~17]{feldman2018privacy} we have
	\begin{align*}
		\lVert\theta_{k+1}-\theta^*\rVert_2^2\leq &\lVert\theta_k-\eta\nabla\Loss_\D(\theta_k)
		+\sqrt{2\eta\sig^2}\mathcal{N}(0,\mathbb{I}_d)-(\theta^*-\eta\nabla\Loss_\D(\theta^*))\rVert_2^2\\
		=&\lVert\theta_k-\theta^*\rVert_2^2+\eta^2\lVert\nabla\Loss_\D(\theta_k)-\nabla\Loss_\D(\theta^*)\rVert_2^2
		+2\eta\sigma^2\lVert\mathcal{N}(0,\mathbb{I}_d)\rVert_2^2\\
		\quad&+2\langle\theta_k-\theta^*,\sqrt{2\eta\sig^2}\mathcal{N}(0,\mathbb{I}_d)\rangle
		-2\eta\langle\nabla\Loss_\D(\theta_k)-\nabla\Loss_\D(\theta^*),\sqrt{2\eta\sig^2}\mathcal{N}(0,\mathbb{I}_d)\rangle\\
		\quad&-2\eta\langle\theta_k-\theta^*,\nabla\Loss_\D(\theta_k)-\nabla\Loss_\D(\theta^*)\rangle.
	\end{align*}
	By $\smh$-smoothness of $\Loss_D$ and $\eta=\frac{\lam}{2\beta^2}$, we have
	\begin{equation}
		\eta^2\lVert\nabla\Loss_\D(\theta_k)-\nabla\Loss_\D(\theta^*)\rVert_2^2\leq \eta\lam\lVert\theta_k-\theta^*\rVert_2^2.
	\end{equation}

	By strong convexity of $\Loss_D$, we have
	\begin{align*}
		\mathbb{E}[\langle\nabla \Loss_D(\theta_k),\theta_k-\theta^*\rangle]&\geq \mathbb{E}[\Loss_D(\theta_k)-\Loss_D(\theta^*)] +\frac{\lambda}{2}\mathbb{E}[\lVert\theta_k-\theta^*\rVert^2]\\
		&\geq \frac{\lambda}{2}\mathbb{E}[\lVert\theta_k-\theta^*\rVert^2]+\frac{\lambda}{2}\mathbb{E}[\lVert\theta_k-\theta^*\rVert^2]\\
		&\geq \lambda\mathbb{E}[\lVert\theta_k-\theta^*\rVert^2].
	\end{align*}


	By taking expectations on the controlling inequality, and plugging the above results, we get
	\begin{equation}
		\label{eqn:distance_recursive}
		\mathbb{E}[\lVert\theta_{k+1}-\theta^*\rVert_2^2]\leq (1-\lam\eta)\mathbb{E}[\lVert\theta_k-\theta^*\rVert_2^2]+2\eta\sigma^2d.
	\end{equation}

	By $\beta$-smoothness,
	\begin{equation*}
		\Loss_\D(\theta_k)-\Loss_\D(\theta^*)\leq \langle\nabla \Loss_\D(\theta^*),\theta_k-\theta^*\rangle+ \frac{\beta}{2}\lVert\theta_{k}-\theta^*\rVert_2^2.
	\end{equation*}

	By the optimality of $\theta^*$ in the relative interior of convex set $\mathcal{C}$ and the fact that $\theta_{K}\in\mathcal{C}$, we prove 
	\begin{equation*}
		\langle\nabla \Loss_\D(\theta^*),\theta_K-\theta^*\rangle=0.
	\end{equation*}
	Therefore, $\Loss_\D(\theta_K)-\Loss_\D(\theta^*)\leq \frac{\beta}{2}\lVert\theta_{K}-\theta^*\rVert_2^2$. On taking expectation over $\theta_K$, we have
	\begin{equation*}
		\mathbb{E}[\Loss_D(\theta_{K})-\Loss_D(\theta^*)]\leq \frac{\beta}{2}\mathbb{E}[\lVert\theta_{K}-\theta^*\rVert^2].
	\end{equation*}
	On unrolling the recursion in~\eqr{eqn:distance_recursive}, we have 
	\begin{align*}
		\mathbb{E}[\Loss_D(\theta_{K})-\Loss_D(\theta^*)]& \leq \frac{\beta}{2}(1-\eta\lambda)^\K\mathbb{E}[\lVert\theta_{0}-\theta^*\rVert_2^2]+{2\beta d\sig^2}\sum_{k=0}^{\K-1}(1-\eta\lambda)^k\\
		\leq &\frac{\beta}{2}e^{-\lambda\eta \K}\mathbb{E}[\lVert\theta_{0}-\theta^*\rVert_2^2]+\frac{2\beta d\sig^2}{\lambda}.
	\end{align*}
	Since we always have $\norm{\C} \leq 2\Lip / \lambda$, we can bound $\mathbb{E}[\lVert\theta_{0}-\theta^*\rVert_2^2]\leq\frac{4\Lip^2}{\lambda^2}$ as both $\ptheta_0, \ptheta^* \in \C$.
	Therefore, we have
	\begin{equation*}
		\mathbb{E}[\Loss_D(\theta_{K})-\Loss_D(\theta^*)]\leq  \frac{2\beta L^2}{\lambda^2}e^{-\lambda\eta K} +\frac{2\beta d\sigma^2}{\lambda}.
	\end{equation*}
\end{proof}

%
%


\end{document}